\PassOptionsToPackage{unicode}{hyperref}
\PassOptionsToPackage{hyphens}{url}
\PassOptionsToPackage{dvipsnames,svgnames,x11names}{xcolor}
\documentclass[
  12pt]{article}

\usepackage{amsmath,amssymb}
\usepackage{iftex}
\ifPDFTeX
  \usepackage[T1]{fontenc}
  \usepackage[utf8]{inputenc}
  \usepackage{textcomp} 
\else 
  \usepackage{unicode-math}
  \defaultfontfeatures{Scale=MatchLowercase}
  \defaultfontfeatures[\rmfamily]{Ligatures=TeX,Scale=1}
\fi
\usepackage{lmodern}
\ifPDFTeX\else  
\fi
\IfFileExists{upquote.sty}{\usepackage{upquote}}{}
\IfFileExists{microtype.sty}{
  \usepackage[]{microtype}
  \UseMicrotypeSet[protrusion]{basicmath} 
}{}
\makeatletter
\@ifundefined{KOMAClassName}{
  \IfFileExists{parskip.sty}{%
    \usepackage{parskip}
  }{
    \setlength{\parindent}{0pt}
    \setlength{\parskip}{6pt plus 2pt minus 1pt}}
}{
  \KOMAoptions{parskip=half}}
\makeatother
\usepackage{xcolor}
\makeatletter
\ifx\paragraph\undefined\else
  \let\oldparagraph\paragraph
  \renewcommand{\paragraph}{
    \@ifstar
      \xxxParagraphStar
      \xxxParagraphNoStar
  }
  \newcommand{\xxxParagraphStar}[1]{\oldparagraph*{#1}\mbox{}}
  \newcommand{\xxxParagraphNoStar}[1]{\oldparagraph{#1}\mbox{}}
\fi
\ifx\subparagraph\undefined\else
  \let\oldsubparagraph\subparagraph
  \renewcommand{\subparagraph}{
    \@ifstar
      \xxxSubParagraphStar
      \xxxSubParagraphNoStar
  }
  \newcommand{\xxxSubParagraphStar}[1]{\oldsubparagraph*{#1}\mbox{}}
  \newcommand{\xxxSubParagraphNoStar}[1]{\oldsubparagraph{#1}\mbox{}}
\fi
\makeatother

\def\calA{\mathcal{A}}

\def\calM{\mathcal{M}}  
 \def\calQ{\mathcal{Q}} 
\def\calS{\mathcal{S}} \def\calT{\mathcal{T}} 
  \def\calX{\mathcal{X}}

\def\E{\mathbb{E}}
\def\P{\mathbb{P}}

\usepackage{amsthm}
\usepackage{algorithm}
\usepackage{algpseudocode}

\newtheorem{assumption}{Assumption}[section]
\newtheorem{lemma}{Lemma}[section]
\newtheorem{theorem}{Theorem}[section]
\newtheorem{proposition}{Proposition}

\usepackage{longtable,booktabs,array}
\usepackage{calc} 
\usepackage{etoolbox}
\makeatletter
\patchcmd\longtable{\par}{\if@noskipsec\mbox{}\fi\par}{}{}
\makeatother
\IfFileExists{footnotehyper.sty}{\usepackage{footnotehyper}}{\usepackage{footnote}}
\makesavenoteenv{longtable}
\usepackage{graphicx}
\makeatletter
\def\maxwidth{\ifdim\Gin@nat@width>\linewidth\linewidth\else\Gin@nat@width\fi}
\def\maxheight{\ifdim\Gin@nat@height>\textheight\textheight\else\Gin@nat@height\fi}
\makeatother
\setkeys{Gin}{width=\maxwidth,height=\maxheight,keepaspectratio}
\makeatletter
\def\fps@figure{htbp}
\makeatother

\makeatletter
\@ifpackageloaded{caption}{}{\usepackage{caption}}
\AtBeginDocument{%
\ifdefined\contentsname
  \renewcommand*\contentsname{Table of contents}
\else
  \newcommand\contentsname{Table of contents}
\fi
\ifdefined\listfigurename
  \renewcommand*\listfigurename{List of Figures}
\else
  \newcommand\listfigurename{List of Figures}
\fi
\ifdefined\listtablename
  \renewcommand*\listtablename{List of Tables}
\else
  \newcommand\listtablename{List of Tables}
\fi
\ifdefined\figurename
  \renewcommand*\figurename{Figure}
\else
  \newcommand\figurename{Figure}
\fi
\ifdefined\tablename
  \renewcommand*\tablename{Table}
\else
  \newcommand\tablename{Table}
\fi
}
\@ifpackageloaded{float}{}{\usepackage{float}}
\floatstyle{ruled}
\@ifundefined{c@chapter}{\newfloat{codelisting}{h}{lop}}{\newfloat{codelisting}{h}{lop}[chapter]}
\floatname{codelisting}{Listing}

\makeatother
\makeatletter
\@ifpackageloaded{caption}{}{\usepackage{caption}}
\@ifpackageloaded{subcaption}{}{\usepackage{subcaption}}
\makeatother

\ifLuaTeX
  \usepackage{selnolig}  
\fi
\usepackage[]{natbib}
\usepackage{bookmark}

\IfFileExists{xurl.sty}{\usepackage{xurl}}{} 
\hypersetup{
  pdftitle={Direct Knowledge Transfer for Offline Q-Learning},
  pdfauthor={Zhanye Luo; Guangyan Gan; Lin William Cong},
  pdfkeywords={Offline reinforcement learning; Transfer learning; Joint estimation; Fitted Q-iteration; Fused regularization},
  colorlinks=true,
  linkcolor={blue},
  filecolor={Maroon},
  citecolor={Blue},
  urlcolor={Blue}}

\begin{document}

\def\spacingset#1{\renewcommand{\baselinestretch}%
{#1}\small\normalsize} \spacingset{1}


\title{\bf Direct Knowledge Transfer\\ for Offline $Q$-Learning}
  \author{%
    Zhanye Luo\thanks{Faculty of Business and Economics, The University of Hong Kong.
      Email: \texttt{zhanye.luo@connect.hku.hk}.}
    \and
    Guangyan Gan\thanks{Nanyang Business School,
      Nanyang Technological University, Singapore.
      Email: \texttt{GUANGYAN001@e.ntu.edu.sg}.}
    \and
    Lin William Cong\thanks{Corresponding author.
      Nanyang Business School, Nanyang Technological University, Singapore.
      Email: \texttt{will.cong@ntu.edu.sg}.}
  }
  \date{}
  \maketitle

\bigskip
\begin{abstract}
Data-driven sequential decision making in marketing and healthcare
can benefit from observations collected in related environments,
particularly when data in the environment of interest are limited.
Transferring such information requires accounting for differences
in rewards and transition dynamics across environments.
We study knowledge transfer for offline reinforcement learning
through a joint estimation framework that directly integrates source
information into target value-function learning.
We propose Direct Transfer Fitted $Q$-Iteration (DFQI),
which updates the target $Q$-function while simultaneously
borrowing the source information through an $\ell_1$ penalty 
that links the target coefficients to a weighted
aggregate of source coefficients, with both determined
jointly within each Bellman update.
Using sieve approximation, we establish statistical guarantees
for target estimation and policy learning in stationary
discounted environments with common or heterogeneous
transition kernels.
Empirical studies further demonstrate the effectiveness
of the proposed DFQI framework in transferring source information to improve
policy learning with limited target data.
\end{abstract}

\noindent%

\section{Introduction}
\label{sec:introduction}

Many sequential decision-making problems call for learning effective
policies from previously collected data. Examples include dynamic
customer targeting in marketing
\citep{Simester2006DynamicMailing}
and adaptive treatment decisions in healthcare
\citep{Murphy2003OptimalDTR,Komorowski2018AIClinician}.
We focus on the offline setting, in which policies are learned
from a fixed dataset without further interaction with the
environment during learning \citep{Levine2020OfflineRL}.
Reinforcement learning (RL) provides a principled framework for studying
such sequential decision problems, with policies evaluated
by their expected cumulative rewards
\citep{Kaelbling1996RLSurvey,
Sutton2018ReinforcementLearning}.

In stationary, discounted Markov decision processes (MDPs),
fitted $Q$-iteration (FQI) provides a standard approach to
offline policy learning
\citep{Ernst2005BatchRL,Riedmiller2005NeuralFQI}.
It approximates the optimal action-value function through
successive Bellman regressions, using observed transitions
and the preceding $Q$ estimate to construct pseudo-responses.
Then, the fitted $Q$-function determines a greedy policy.
From a statistical perspective, these updates involve estimating
regression functions from a finite dataset of transitions
\citep{Murphy2005QGeneralization, Clifton2020QLearningReview}.
Limited target data can therefore constrain estimation accuracy,
while collecting additional trajectories may be costly or
impractical \citep{Levine2020OfflineRL}.
Because successive regressions use estimated pseudo-responses,
errors in one update can propagate to later updates and affect
the value of the learned policy
\citep{Munos2008FittedValueIteration,
Farahmand2010ErrorPropagation,Fan2020DeepQLearning}.
These considerations motivate seeking additional information
to improve estimation of the target $Q$-function.

Transfer learning offers a way to address this limitation by using
information from related source tasks to improve learning in the
target task \citep{Taylor2009TransferRLSurvey, Pan2010TransferLearningSurvey}. 
In many applications, such information is available in historical records from related populations or operating environments, but difficult to obtain from a small target sample
alone. Statistical research on transfer learning has clarified when
and to what extent related source data can improve target
learning. In regression settings, theoretical guarantees
characterize how these gains depend on source and target
sample sizes and the similarity between tasks
\citep{Li2022TransLasso,Tian2023TransGLM}.
In transfer $Q$-learning, source data describe environments
that may differ from the target in rewards, transition dynamics,
and the distributions of observed states and actions.
Accurate knowledge of a source environment does not, by itself,
ensure improved policy learning in the target
\citep{Chen2026BatchQLearning}.
The challenge is therefore to use shared information to improve
estimation of the target $Q$-function while accounting for
differences relevant to the target policy.

A prominent strategy in statistical transfer learning follows
a two-step design: a source-informed reference is first
estimated, and target observations are then used to estimate
a correction
\citep{Bastani2021PredictingWithProxies,Li2022TransLasso}.
Although these methods provide theoretical guarantees,
the reference remains fixed during the correction step,
leaving any remaining mismatch with the target to be
addressed through the correction alone.
When task discrepancies are substantial, the reference may
require a correction that is difficult to estimate accurately
from limited target data
\citep{Tian2023TransGLM}.
In FQI-based transfer learning, this concern extends beyond
an individual regression: errors remaining after correction
can propagate through subsequent Bellman responses and affect
the policy ultimately learned
\citep{Farahmand2010ErrorPropagation,
Chen2026BatchQLearning}.
These considerations motivate a statistical framework for
transfer $Q$-learning that jointly determines a source-informed
reference and its target-specific correction within each
Bellman update.

We introduce Direct Transfer Fitted $Q$-Iteration (DFQI; Algorithm~\ref{alg:joint-transfer-fqi}),
a new statistical framework for incorporating source
information into target $Q$-function estimation.
Its defining feature is to integrate information borrowing
and target adaptation within a single joint optimization problem
at each Bellman update.
Each task constructs Bellman pseudo-responses from its own
preceding $Q$ estimate.
Under a common sieve representation, the joint objective
combines the target and auxiliary source regression losses
with an $\ell_1$ penalty on the difference between the target
coefficients and a sample-size-weighted aggregate of source
coefficients.
The penalty encourages a sparse target correction to the
source aggregate while retaining task-specific fitted functions.
The resulting target $Q$-function determines the final
greedy policy.

For theoretical analysis, we consider the sieve
implementation of DFQI, which approximates
nonparametric $Q$-functions through finite-dimensional
basis expansions
\citep{Shi2022ValueInference,Chen2026BatchQLearning}.
Under suitable regularity and transfer conditions, we first
derive target estimation bounds for tasks with common
transition kernels but potentially different reward functions.
The bounds distinguish sieve approximation error,
a statistical component governed by the combined source
and target sample size, and a target-correction component
depending on the available target data and the magnitude
and sparsity of the aggregate discrepancy.
We then extend the analysis to tasks with heterogeneous
transition kernels under additional stability conditions.
The resulting bounds retain the same structure, with the
correction term accounting for discrepancies in both rewards
and transition dynamics.
In both settings, we trace estimation errors through the
Bellman recursion to obtain value-loss bounds for the learned
target policy, also accounting for the finite number of
FQI iterations.

To complement the theoretical analysis, we conduct
Monte Carlo simulations to evaluate target $Q$-function
estimation and a semisynthetic study based on real data
to evaluate policy performance.
The results provide empirical support for the effectiveness
of DFQI with limited target data in the settings
considered.

The remainder of this paper is organized as follows.
Section~\ref{sec:related-works} reviews the related literature.
Section~\ref{sec:methodology} formulates the transfer learning
problem, and Section~\ref{sec:joint-fqi} presents the DFQI
framework and its sieve implementation.
Section~\ref{sec:theory} establishes statistical guarantees
under common and heterogeneous transition kernels.
Section~\ref{sec:empirical} presents the empirical studies,
and Section~\ref{sec:conclusion} concludes.
Proofs are provided in Appendix~\ref{app:proofs}.

\section{Literature Review}
\label{sec:related-works}
Our study draws on two strands of literature: offline reinforcement
learning and knowledge transfer. We first review statistical
approaches to offline policy learning, then discuss transfer
methods in regression and reinforcement learning.

\paragraph{Offline reinforcement learning.}
Regression-based approaches provide a statistical foundation
for learning policies from offline data, with fitted
$Q$-iteration serving as a basic algorithmic framework
\citep{Ernst2005BatchRL}.
Analyses of fitted value iteration and $Q$-learning relate
function approximation and finite-sample estimation to policy
performance, and examine how estimation errors propagate
through successive updates
\citep{Murphy2005QGeneralization,Munos2008FittedValueIteration,
Farahmand2010ErrorPropagation,Fan2020DeepQLearning}.
This literature provides tools for studying both individual
Bellman regressions and the policies obtained from the
resulting sequence of fitted functions.

A related line investigates the data coverage and function-class
conditions required for statistically efficient offline learning
\citep{Chen2019BatchRLTheory,Xie2021BVFT,Foster2022OfflineBarriers}.
These studies clarify how learning guarantees depend on the
interaction between the observed distribution and the
representation of value functions.
Pessimistic methods further incorporate uncertainty into
value estimation and policy optimization, providing guarantees
when offline data offer limited support for some decisions
\citep{Jin2021PessimismOfflineRL,Shi2022PessimisticQLearning}.

Statistical approaches to long-term decision making also
study policy optimization over prespecified policy classes
under discounted- and average-reward criteria
\citep{Luckett2020VLearning,Liao2022BatchAverageReward}.
Complementary research develops methods for evaluating
long-term policy values and conducting statistical inference
in infinite-horizon settings
\citep{Shi2022ValueInference,Kallus2022InfiniteHorizonDRL}.
These contributions address policy construction, value
estimation, and uncertainty quantification within an environment.
Our study builds on these statistical foundations to examine
how information from related environments can improve
target $Q$-function estimation and policy learning.

\paragraph{Transfer learning.}
Statistical transfer learning studies the value of auxiliary
data under explicit relationships between source and target
distributions. Minimax analyses of nonparametric classification
and regression characterize how task relatedness and sample
sizes determine attainable target accuracy
\citep{Cai2021TransferClassification,Cai2024TransferRegression}.
In regression, a common principle is to borrow shared
information while accounting for target-specific differences.
Methods using
proxy data, sparse linear models, and generalized linear models
formalize this principle and establish conditions under which
transfer improves target estimation
\citep{Bastani2021PredictingWithProxies,Li2022TransLasso,
Tian2023TransGLM}.
These studies motivate borrowing information through structured
target adaptation, with transfer gains depending on both the
information supplied by source data and the complexity of the
adjustment required for the target.

Related developments address task heterogeneity through
constrained estimation and fused regularization
\citep{Li2024TransHDGLM,He2024TransFusion,He2026TransMission}.
These constructions differ in their treatment of covariate shifts,
regularization, and target adaptation; joint fitting can also
precede a separate correction, as in TransFusion.
This literature provides a foundation for borrowing information
through joint estimation. Extending such constructions to FQI
requires analyzing regression responses that depend on previously
estimated value functions, as well as the policy learned
through the resulting recursion.

\paragraph{Transfer learning for reinforcement learning.}
Within RL, transfer methods differ in both the knowledge
transferred and the data available for adaptation
\citep{Taylor2009TransferRLSurvey,Zhu2023DeepTransferRLSurvey}.
Successor features separate reward dependence from transition
dynamics, enabling reuse of value information across tasks
with common dynamics and different rewards
\citep{Barreto2017SuccessorFeatures}.
Representation-based methods learn shared features from
source tasks to improve downstream learning
\citep{Agarwal2023RepresentationalTransfer},
and such benefits have also been established when both source
and target learning use offline data
\citep{Ishfaq2024OfflineRepresentation}.
Other approaches exploit shared structure in transition
models \citep{Chai2025CompositeMDP} or combine offline
source observations with online target interaction when
source and target dynamics differ
\citep{Qu2025HybridTransferRL}.
Our focus is on transferring information at the level of
value-function regression using fixed offline datasets,
with the target policy as the object of interest.

More directly related are methods that transfer information
within Bellman regressions. Importance-weighted transfer FQI adjusts the contribution
of source samples to account for differences in rewards
and transition dynamics
\citep{Tirinzoni2018ImportanceWeightedTransfer}.
Finite-horizon approaches combine
backward regression with target-oriented response construction,
including neural methods that reweight source transitions
to accommodate differences in dynamics
\citep{Chen2025FiniteHorizonTransfer,Chai2025DeepTransferQLearning}.
The closest work is \citet{Chen2026BatchQLearning}, which studies
transfer FQI in stationary discounted environments.
At each iteration, it forms task-specific Bellman responses,
fits a pooled reference using target and source observations,
and then estimates task-specific corrections.
Its theory relates sample sizes and task discrepancies to
learning accuracy while accounting for Bellman error propagation.
Our contribution concerns the construction and estimation of
the source-informed reference within each Bellman update.
DFQI jointly estimates the target and task-specific source
functions, using a weighted aggregate of the source functions
as its reference.
We establish statistical guarantees for this joint construction,
characterizing how information borrowing and target correction
affect target estimation and policy learning through the
Bellman recursion.

\section{Methodology}
\label{sec:methodology}

\subsection{Statistical Formulation of Reinforcement Learning}

We consider an infinite-horizon, time-homogeneous discounted
Markov decision process (MDP) \citep{Sutton2018ReinforcementLearning},
specified by
\[
\calM=\{\calX,\calA,P,r,\gamma,\nu\}.
\]
Here, $\calX\subseteq\mathbb R^d$ is the state space,
with $d$ denoting the state dimension; $\calA$ is a
finite action space with $m$ actions, $\gamma\in[0,1)$
is the discount factor, and $\nu$ is the initial-state
distribution.
At each time $t$, the agent observes $X_t\in\calX$, selects
$A_t\in\calA$, receives a reward $R_t$, and transitions to
$X_{t+1}$.
Conditional on the current state-action pair, the distribution
of the next state is independent of the preceding history
and is determined by the transition kernel $P$:
\[
\P(X_{t+1}\in B\mid X_t=x,A_t=a)
=
P(B\mid x,a),
\qquad B\subseteq\calX \text{ measurable}.
\]
The conditional mean reward is
$r(x,a)=\E[R_t\mid X_t=x,A_t=a]$.

A stationary Markov policy $\pi$ selects action $a$ in state
$x$ with probability $\pi(a\mid x)$.
For a deterministic policy, $\pi(x)$ denotes the action
selected with probability one at state $x$.
Writing $\E_\pi$ for expectation under policy $\pi$,
we define the state-value and action-value functions by
\[
\begin{aligned}
V^\pi(x)
&=
\E_\pi\left[
\sum_{t=0}^{\infty}\gamma^t R_t
\,\middle|\,
X_0=x
\right],\\
Q^\pi(x,a)
&=
\E_\pi\left[
\sum_{t=0}^{\infty}\gamma^t R_t
\,\middle|\,
X_0=x,\ A_0=a
\right].
\end{aligned}
\]
In $Q^\pi(x,a)$, the initial action is fixed at $a$,
and policy $\pi$ is followed thereafter.

The optimal action-value function is
\[
Q^*(x,a)=\sup_\pi Q^\pi(x,a),
\]
and satisfies the Bellman optimality equation
\[
Q^*(x,a)
=
\E\left[
R_t+\gamma\max_{a'\in\calA}Q^*(X_{t+1},a')
\,\middle|\,
X_t=x,\ A_t=a
\right].
\]
Equivalently,
\[
\E\left[
R_t+\gamma\max_{a'\in\calA}Q^*(X_{t+1},a')
-Q^*(X_t,A_t)
\,\middle|\,
X_t,A_t
\right]=0.
\]
This conditional moment equation states that the Bellman
residual has conditional mean zero, providing a basis for
regression-based estimation of $Q^*$.

Our goal is to improve estimation of $Q^*$ for a target task
by transferring information from related source tasks.
Estimation of $Q^*$ provides a basis for policy learning,
since an optimal deterministic policy can be obtained as
\[
\pi^*(x)\in\arg\max_{a\in\calA}Q^*(x,a).
\]
In practice, we construct a greedy policy by replacing
$Q^*$ with its estimate.

\subsection{Target and Source Data}

We consider an offline transfer reinforcement learning
setting with one target task and $K$ source tasks.
The target task is indexed by $k=0$, and the source tasks
by $k\in[K]=\{1,\ldots,K\}$.
All tasks share $\calX$, $\calA$, and $\gamma$, but may
differ in their transition kernels, mean reward functions,
and initial-state distributions:
\[
\calM^{(k)}
=
\{\calX,\calA,P^{(k)},r^{(k)},\gamma,\nu^{(k)}\},
\qquad k\in\{0\}\cup[K].
\]
Let $Q^{*(k)}$ denote the optimal action-value function
of task $k$.
For each task $k$, we observe $I^{(k)}$ independent
trajectories, each containing $T$ transitions, collected
under a behavior policy $b^{(k)}$.
The resulting offline dataset is
\[
\calS^{(k)}
=
\left\{
\left(
X_{i,t}^{(k)},
A_{i,t}^{(k)},
R_{i,t}^{(k)},
X_{i,t+1}^{(k)}
\right)
:
1\le i\le I^{(k)},\ 0\le t\le T-1
\right\},
\]
where $i$ indexes trajectories and $t$ indexes decision times.
Thus, task $k$ contributes $I^{(k)}T$ transition observations,
which may be temporally dependent within each trajectory.
The behavior policy satisfies
\[
\P\!\left(
A_{i,t}^{(k)}=a
\mid X_{i,t}^{(k)}=x
\right)
=
b^{(k)}(a\mid x).
\]
The target dataset $\calS^{(0)}$ is typically limited,
while the source datasets $\{\calS^{(k)}\}_{k=1}^K$
provide additional data from related tasks.

\subsection{Similarity Between Target and Source Tasks}

Following \citet{Chen2026BatchQLearning}, we characterize
task discrepancies through differences in mean reward
functions and transition kernels.
For completeness, we also record the Bellman decomposition
used in their task-similarity analysis.

Let $\rho^{(k)}(x'\mid x,a)$ denote the density of the
transition kernel $P^{(k)}(\cdot\mid x,a)$.
For each source task $k\in[K]$, define the reward and
transition discrepancies relative to the target task by
\[
\begin{aligned}
\delta_r^{(k)}(x,a)
&=
r^{(k)}(x,a)-r^{(0)}(x,a),\\
\delta_\rho^{(k)}(x'\mid x,a)
&=
\rho^{(k)}(x'\mid x,a)-\rho^{(0)}(x'\mid x,a).
\end{aligned}
\]

\begin{lemma}[Decomposition of action-value discrepancy]
\label{lem:q-discrepancy-decomposition}
For any source task $k\in[K]$, define
\[
\delta_Q^{(k)}(x,a)
:=
Q^{*(k)}(x,a)-Q^{*(0)}(x,a).
\]
Then
\[
\begin{aligned}
\delta_Q^{(k)}(x,a)
={}&
\delta_r^{(k)}(x,a)\\
&+
\gamma
\int_{\calX}
\max_{a'\in\calA} Q^{*(k)}(x',a')
\,\delta_\rho^{(k)}(x'\mid x,a)\,dx'\\
&+
\gamma
\int_{\calX}
\left[
\max_{a'\in\calA}Q^{*(k)}(x',a')
-
\max_{a'\in\calA}Q^{*(0)}(x',a')
\right]
\rho^{(0)}(x'\mid x,a)\,dx'.
\end{aligned}
\]
\end{lemma}

A short derivation is provided in the appendix.
Lemma~\ref{lem:q-discrepancy-decomposition} relates
differences in optimal action-value functions to reward
and transition discrepancies and their propagated effects
on future values.
This motivates borrowing information across related tasks
while accommodating differences in their $Q$-functions.
In the joint estimation procedure below, we implement
transfer by regularizing the difference between the target
candidate $Q$-function and a weighted aggregate of the
source candidate $Q$-functions at each iteration.

\section{Direct Q-Learning for Data Transfer}
\label{sec:joint-fqi}
\subsection{Direct Transfer Fitted Q-Iteration}

The proposed joint transfer algorithm builds on fitted
$Q$-iteration for learning optimal action-value functions
from offline data \citep{Ernst2005BatchRL}.
At each iteration, we construct each task's Bellman
pseudo-responses using its own preceding $Q$-function
estimate.
The target and source functions are then jointly updated
by minimizing their task-specific regression losses
together with a penalty on the target correction relative
to a weighted source aggregate.
The source-informed reference and the target correction
are thus determined within the same optimization.

For theoretical analysis, we partition the trajectories
of each task into $\Upsilon$ disjoint groups of equal size,
one for each iteration.
For simplicity, assume that $I^{(k)}$ is divisible by
$\Upsilon$ for every task.
Let $\calS_\tau^{(k)}$ contain all transitions from the
trajectories in group $\tau$, used at iteration $\tau$.
The number of transition observations from task $k$
at each iteration is therefore
\[
n_k
=
|\calS_\tau^{(k)}|
=
\frac{I^{(k)}T}{\Upsilon},
\qquad k\in\{0\}\cup[K].
\]
Here and below, $(i,t)\in\calS_\tau^{(k)}$ means that
the transition indexed by $(i,t)$ belongs to
$\calS_\tau^{(k)}$.
Define the total source sample size and source weights by
\[
n_s=\sum_{k=1}^K n_k,
\qquad
w_k=\frac{n_k}{n_s},
\qquad k\in[K].
\]
These source weights remain fixed across iterations.
At iteration $\tau$, given the preceding estimates
$\{\widehat Q_{\tau-1}^{(k)}\}_{k=0}^K$, we construct
the pseudo-response for each task $k\in\{0\}\cup[K]$ as
\begin{equation}
\label{eq:bellman-response}
\widehat Y_{i,t}^{(k),\tau}
=
R_{i,t}^{(k)}
+
\gamma
\max_{a'\in\calA}
\widehat Q_{\tau-1}^{(k)}
\left(X_{i,t+1}^{(k)},a'\right),
\qquad (i,t)\in\calS_\tau^{(k)}.
\end{equation}
Let $\calT^{(k)}$ denote the Bellman optimality operator
for task $k$, defined by
\[
(\calT^{(k)}Q)(x,a)
=
r^{(k)}(x,a)
+
\gamma
\int_{\calX}
\max_{a'\in\calA}Q(x',a')
\,\rho^{(k)}(x'\mid x,a)\,dx'.
\]
Conditional on the preceding fitted functions,
the corresponding population regression function is
$f_\tau^{(k)}
=
\calT^{(k)}\widehat Q_{\tau-1}^{(k)}. $
These pseudo-responses are held fixed during the current
joint regression update.
Let $\calQ$ be a class of candidate action-value functions.
For each task $k\in\{0\}\cup[K]$ and $Q\in\calQ$, define
the empirical Bellman regression loss
\[
\mathcal L_\tau^{(k)}(Q)
=
\frac{1}{2n_k}
\sum_{(i,t)\in\calS_\tau^{(k)}}
\left[
\widehat Y_{i,t}^{(k),\tau}
-
Q\left(X_{i,t}^{(k)},A_{i,t}^{(k)}\right)
\right]^2.
\]
Write $\mathbf Q=(Q^{(0)},\ldots,Q^{(K)})$ and define
the source reference by
$\overline Q_s
=
\sum_{k=1}^K w_k Q^{(k)}.$
Let $\mathcal R_{\Delta,\tau}(\mathbf Q)$ be a regularizer
for the target correction $Q^{(0)}-\overline Q_s$.
The joint update is
\begin{equation}
\label{eq:joint-fqi-update}
\widehat{\mathbf Q}_\tau
\in
\operatorname*{arg\,min}_{\mathbf Q\in\calQ^{K+1}}
\Bigl\{
\mathcal L_\tau^{(0)}(Q^{(0)})
+
\mu\sum_{k=1}^K w_k
\mathcal L_\tau^{(k)}(Q^{(k)})
+
\lambda_\Delta
\mathcal R_{\Delta,\tau}(\mathbf Q)
\Bigr\},
\end{equation}
where
$\widehat{\mathbf Q}_\tau
=(\widehat Q_\tau^{(0)},\ldots,\widehat Q_\tau^{(K)})$.
Here, $\mu\ge0$ controls the contribution of the source
regression losses, and $\lambda_\Delta\ge0$ controls
the strength of transfer regularization.
The source regression losses keep the source functions
tied to their own Bellman pseudo-responses.
Through the transfer penalty, the target regression loss
can influence the source aggregate within the same update,
allowing the reference to adjust as the target correction
is estimated.
Section~\ref{subsec:sieve-approximation} develops
a sieve implementation of the joint update,
in which the regularizer is an $\ell_1$ penalty
on the correction coefficients.
After $\Upsilon$ iterations, we use
$\widehat Q_\Upsilon^{(0)}$ to estimate the target optimal
action-value function $Q^{*(0)}$ and construct the
corresponding greedy policy
\[
\widehat\pi_\Upsilon^{(0)}(x)
\in
\operatorname*{arg\,max}_{a\in\calA}
\widehat Q_\Upsilon^{(0)}(x,a).
\]
Algorithm~\ref{alg:joint-transfer-fqi} summarizes
the procedure.

\subsection{Sieve Function Approximation for Direct Transfer FQI}
\label{subsec:sieve-approximation}

We implement Algorithm~\ref{alg:joint-transfer-fqi}
using a common sieve space for all tasks.
By representing candidate functions as linear combinations
of basis functions whose number can grow with the sample
size, sieve approximation combines flexible function
approximation with finite-dimensional estimation
\citep{Chen2007Sieve}.
In our joint update, the common basis allows the
weighted source reference and the target's deviation
from it to be expressed directly in terms of the
task-specific coefficients.
With the Bellman pseudo-responses held fixed, the squared
regression losses and the $\ell_1$ correction penalty
yield a convex optimization problem over the task-specific
coefficients.

Let $\phi(x)=(\phi_1(x),\ldots,\phi_p(x))^\top$
collect $p$ prespecified sieve basis functions on $\calX$.
For notational convenience, label the actions by
$1,\ldots,m$.
Define the action-specific feature vector
$\xi(x,a)
=
\left[
\phi^\top(x)\mathbf 1(a=1),
\ldots,
\phi^\top(x)\mathbf 1(a=m)
\right]^\top
\in\mathbb R^q,
 q=mp, $
where $\mathbf 1(\cdot)$ denotes the indicator function.
In the sieve implementation, we take $\calQ=\calQ_p$,
where
\[
\calQ_p
=
\left\{
(x,a)\mapsto\xi^\top(x,a)\beta:
\beta\in\mathbb R^q
\right\}.
\]
For each task $k\in\{0\}\cup[K]$, a candidate
action-value function therefore has the form
\[
Q^{(k)}(x,a)
=
\xi^\top(x,a)\beta^{(k)},
\qquad
\beta^{(k)}\in\mathbb R^q.
\]

At iteration $\tau$, for each task $k\in\{0\}\cup[K]$,
let $Z_\tau^{(k)}\in\mathbb R^{n_k\times q}$
denote the design matrix with rows
\[
\xi^\top\left(X_{i,t}^{(k)},A_{i,t}^{(k)}\right),
\qquad (i,t)\in\calS_\tau^{(k)}.
\]
Stack the corresponding pseudo-responses in the same
order to form
\[
\widehat y_\tau^{(k)}
=
\left(
\widehat Y_{i,t}^{(k),\tau}
\right)_{(i,t)\in\calS_\tau^{(k)}}
\in\mathbb R^{n_k}.
\]

Under the common sieve representation, the source
aggregate and target correction satisfy
\[
\begin{aligned}
\overline Q_s(x,a)
&=\xi^\top(x,a)\overline\beta_s,
&\qquad
\overline\beta_s
&=\sum_{k=1}^K w_k\beta^{(k)},\\
Q^{(0)}(x,a)-\overline Q_s(x,a)
&=\xi^\top(x,a)\delta^{(0)},
&\qquad
\delta^{(0)}
&=\beta^{(0)}-\overline\beta_s.
\end{aligned}
\]
We specify the transfer regularizer as
$\mathcal R_{\Delta,\tau}(\mathbf Q)
=
\|\delta^{(0)}\|_1.$
Stack the task-specific coefficients as
\[
\theta
=
\left(
(\beta^{(0)})^\top,\ldots,(\beta^{(K)})^\top
\right)^\top
\in\mathbb R^{(K+1)q}.
\]
The joint coefficient update at iteration $\tau$ is
\begin{equation}
\label{eq:joint-sieve-update}
\begin{aligned}
\widehat\theta_\tau
\in
\operatorname*{arg\,min}_{\theta\in\mathbb R^{(K+1)q}}
\Biggl\{&
\frac{1}{2n_0}
\left\|
\widehat y_\tau^{(0)}
-
Z_\tau^{(0)}\beta^{(0)}
\right\|_2^2
\\
&+
\mu\sum_{k=1}^K\frac{w_k}{2n_k}
\left\|
\widehat y_\tau^{(k)}
-
Z_\tau^{(k)}\beta^{(k)}
\right\|_2^2
\\
&+
\lambda_\Delta
\left\|
\beta^{(0)}
-
\sum_{k=1}^K w_k\beta^{(k)}
\right\|_1
\Biggr\}.
\end{aligned}
\end{equation}
Let $\widehat\beta_\tau^{(k)}$ denote the coefficient
block of $\widehat\theta_\tau$ corresponding to task $k$.
The updated action-value estimates are
\[
\widehat Q_\tau^{(k)}(x,a)
=
\xi^\top(x,a)\widehat\beta_\tau^{(k)},
\qquad k\in\{0\}\cup[K].
\]

For $\tau<\Upsilon$, these updated functions are used
to construct the task-specific Bellman pseudo-responses
at iteration $\tau+1$.

\begin{algorithm}[H]
\caption{Direct Transfer Fitted Q-Iteration}
\label{alg:joint-transfer-fqi}
\begin{algorithmic}[1]
\setlength{\abovedisplayskip}{2pt plus 1pt minus 1pt}
\setlength{\belowdisplayskip}{2pt plus 1pt minus 1pt}
\setlength{\abovedisplayshortskip}{2pt plus 1pt minus 1pt}
\setlength{\belowdisplayshortskip}{2pt plus 1pt minus 1pt}
\setlength{\jot}{1pt}
\Require
Target dataset $\calS^{(0)}$ and source datasets
$\{\calS^{(k)}\}_{k=1}^K$;
function class $\calQ$ and initial estimates
$\{\widehat Q_0^{(k)}\}_{k=0}^K\subset\calQ$;
discount factor $\gamma$ and number of iterations $\Upsilon$;
transfer regularizers
$\{\mathcal R_{\Delta,\tau}\}_{\tau=1}^{\Upsilon}$;
parameters $\mu\ge0$ and $\lambda_\Delta\ge0$.

\Ensure
Target estimator $\widehat Q_\Upsilon^{(0)}$ and
greedy policy $\widehat\pi_\Upsilon^{(0)}$.

\For{$k=0,1,\ldots,K$}
    \State Partition the trajectories in $\calS^{(k)}$
    into $\Upsilon$ disjoint groups of equal size.
    Let $\calS_\tau^{(k)}$ contain all transitions
    from the trajectories in group $\tau$.
    \State Set $n_k=I^{(k)}T/\Upsilon$.
\EndFor

\State Set $n_s=\sum_{k=1}^K n_k$ and
$w_k=n_k/n_s$ for $k\in[K]$.

\Statex For $\mathbf Q=(Q^{(0)},\ldots,Q^{(K)})$,
the regularizer $\mathcal R_{\Delta,\tau}(\mathbf Q)$
penalizes the target correction
$Q^{(0)}-\sum_{k=1}^K w_kQ^{(k)}$.

\For{$\tau=1,2,\ldots,\Upsilon$}
    \For{$k=0,1,\ldots,K$}
        \State Construct the pseudo-responses
        \[
        \widehat Y_{i,t}^{(k),\tau}
        =
        R_{i,t}^{(k)}
        +
        \gamma\max_{a'\in\calA}
        \widehat Q_{\tau-1}^{(k)}
        \left(X_{i,t+1}^{(k)},a'\right),
        \qquad (i,t)\in\calS_\tau^{(k)}.
        \]
    \EndFor

    \State Jointly update
    $\widehat{\mathbf Q}_\tau
    =(\widehat Q_\tau^{(0)},\ldots,\widehat Q_\tau^{(K)})$
    by solving
    \[
    \begin{aligned}
    \widehat{\mathbf Q}_\tau
    \in
    \operatorname*{arg\,min}_{\mathbf Q\in\calQ^{K+1}}
    \Biggl\{&
    \frac{1}{2n_0}
    \sum_{(i,t)\in\calS_\tau^{(0)}}
    \left[
    \widehat Y_{i,t}^{(0),\tau}
    -
    Q^{(0)}\left(X_{i,t}^{(0)},A_{i,t}^{(0)}\right)
    \right]^2
    \\
    &+
    \mu\sum_{k=1}^K\frac{w_k}{2n_k}
    \sum_{(i,t)\in\calS_\tau^{(k)}}
    \left[
    \widehat Y_{i,t}^{(k),\tau}
    -
    Q^{(k)}\left(X_{i,t}^{(k)},A_{i,t}^{(k)}\right)
    \right]^2
    \\
    &+
    \lambda_\Delta
    \mathcal R_{\Delta,\tau}(\mathbf Q)
    \Biggr\}.
    \end{aligned}
    \]
\EndFor

\State Construct the greedy policy
\[
\widehat\pi_\Upsilon^{(0)}(x)
\in
\operatorname*{arg\,max}_{a\in\calA}
\widehat Q_\Upsilon^{(0)}(x,a).
\]
\end{algorithmic}
\end{algorithm}

\begin{proposition}[Convexity and solvability of the joint sieve update]
\label{prop:convexity-solvability}
Fix an iteration $\tau$ and condition on the preceding
fitted functions.
The joint update in \eqref{eq:joint-sieve-update}
can be written as a generalized Lasso problem
\citep{TibshiraniTaylor2011GeneralizedLasso}:
\[
\widehat\theta_\tau
\in
\operatorname*{arg\,min}_{\theta\in\mathbb R^{(K+1)q}}
\left\{
\frac12
\left\|
\widetilde y_\tau-\widetilde Z_\tau\theta
\right\|_2^2
+
\lambda_\Delta\|D\theta\|_1
\right\}.
\]
Here, $\widetilde y_\tau$ stacks the task-specific
pseudo-response vectors scaled to incorporate the
regression weights, and $\widetilde Z_\tau$ is the
correspondingly scaled block-diagonal design matrix.
The contrast matrix $D$ maps the stacked coefficients
to the target correction, so that
$D\theta=\delta^{(0)}$.
The objective is convex and admits at least one minimizer.
\end{proposition}

For brevity, the explicit constructions of
$\widetilde y_\tau$, $\widetilde Z_\tau$, and $D$,
together with the proof, are provided in
Appendix~\ref{app:proof-convexity-solvability}.

\section{Theoretical Results}
\label{sec:theory}

In this section, we study the statistical properties of DFQI
(Algorithm~\ref{alg:joint-transfer-fqi}) under sieve approximation.
We first consider tasks with a common
transition kernel and then extend the analysis to heterogeneous
transition kernels under additional conditions. We begin by
introducing the regularity conditions shared by both settings.

\subsection{Common Regularity Conditions}

\begin{assumption}
\label{assump:data-process}
The following conditions hold for the trajectory data.
\begin{enumerate}
\renewcommand{\labelenumi}{(\roman{enumi})}
\renewcommand{\theenumi}{\theassumption(\roman{enumi})}

\item \textbf{Trajectory splitting.}
Complete trajectories are assigned to iterations independently
of their observed values. Let
$\mathcal D_\tau :=(\calS_\tau^{(0)},\ldots,\calS_\tau^{(K)})$
be the joint data block used at iteration $\tau$.
The blocks $\mathcal D_1,\ldots,\mathcal D_\Upsilon$
are mutually independent. The initial fitted functions
$\{\widehat Q_0^{(k)}\}_{k=0}^K$ are fixed or constructed
from an auxiliary sample independent of these blocks.

\item \textbf{Time-homogeneous dynamics.}
\label{assump:time-homogeneous}
For each task $k\in\{0\}\cup[K]$, conditional on
$(X_{i,t}^{(k)},A_{i,t}^{(k)})=(x,a)$, the next state
follows $P^{(k)}(\cdot\mid x,a)$, with density
$\rho^{(k)}(\cdot\mid x,a)$ almost everywhere.
The kernel does not depend on $i$ or $t$.
The logged data follow a stationary behavior policy
$b^{(k)}$ satisfying
$\mathbb P\!\left( A_{i,t}^{(k)}=a\mid X_{i,t}^{(k)}=x \right) = b^{(k)}(a\mid x)$
for all $i,t,k$.

\item \textbf{Stability and coverage.}
\label{assump:ergodicity}
For each task $k\in\{0\}\cup[K]$, the Markov chain
$\{X_{i,t}^{(k)}\}_{t=0}^{T-1}$ has a unique stationary
distribution with density $\mu^{(k)}$ almost everywhere
and is geometrically ergodic as $T\to\infty$.
Write $\nu^{(k)}(x)$ for the density of the initial distribution $\nu^{(k)}$.
There exist constants $0<c_\mu<C_\mu<\infty$ such that
$c_\mu \le \mu^{(k)}(x),\nu^{(k)}(x) \le C_\mu$ for all $x\in\calX$ and $k\in\{0\}\cup[K]$.

\item \textbf{Reward innovations.}
\label{assump:reward-noise}
For each task $k\in\{0\}\cup[K]$, write
$R_{i,t}^{(k)} = r^{(k)}(X_{i,t}^{(k)},A_{i,t}^{(k)}) +\eta_{i,t}^{(k)}$.
Let $\mathcal H_{i,t}^{(k)}$ be generated by the initial
state, observations through time $t-1$, and the current
pair $(X_{i,t}^{(k)},A_{i,t}^{(k)})$, immediately before
$(R_{i,t}^{(k)},X_{i,t+1}^{(k)})$ is observed.
For every $u\in\mathbb R$,
\[
\begin{aligned}
\mathbb E[
\eta_{i,t}^{(k)}
\mid\mathcal H_{i,t}^{(k)}
]
&=0,\\
\mathbb E\!\left[
\exp\{u\eta_{i,t}^{(k)}\}
\mid\mathcal H_{i,t}^{(k)}
\right]
&\le \exp\{\sigma_\eta^2u^2/2\},
\end{aligned}
\]
where $\sigma_\eta$ is uniform in $i,t,k$.
The Markov property holds relative to the same history:
for every bounded measurable $f$,
\[
\mathbb E[
f(X_{i,t+1}^{(k)})
\mid\mathcal H_{i,t}^{(k)}
]
=
\int_{\calX} f(x')
\rho^{(k)}(x'\mid X_{i,t}^{(k)},A_{i,t}^{(k)})\,dx'.
\]

\end{enumerate}
\end{assumption}

Condition (i) allows the current regression to be analyzed
conditional on preceding fitted functions. Conditions (ii)
and (iv) identify the Bellman conditional mean and control
the residual fluctuations when those fitted functions are
bounded. The density bounds in (iii) enter the norm
comparisons used in the policy-value analysis.

\begin{assumption}
\label{assump:sieve-design}
The following conditions hold for the sieve approximation
and regression design.
\begin{enumerate}
\renewcommand{\labelenumi}{(\roman{enumi})}
\renewcommand{\theenumi}{\theassumption(\roman{enumi})}

\item \textbf{Smoothness, bounded rewards, and approximation.}
\label{assump:smoothness}
For $\kappa,c>0$, let $\ell_\kappa=\lceil\kappa\rceil-1$.
Denote by $\Lambda(\kappa,c)$ the class of functions on
$\calX$ whose partial derivatives up to order $\ell_\kappa$ are
uniformly bounded by $c$ and whose order-$\ell_\kappa$ derivatives
are $(\kappa-\ell_\kappa)$-H\"older continuous with constant at most
$c$ under the Euclidean metric.
There exist $\kappa,c>0$ such that, for every
$k\in\{0\}\cup[K]$, $a\in\calA$, and $x'\in\calX$,
$r^{(k)}(\cdot,a)\in\Lambda(\kappa,c)$ and $\rho^{(k)}(x'\mid\cdot,a)\in\Lambda(\kappa,c)$.
The reward functions satisfy
\[
\sup_{\substack{x\in\calX,\ a\in\calA,\\
k\in\{0\}\cup[K]}}
|r^{(k)}(x,a)|
\le R_{\max}<\infty.
\]
There is a fixed deterministic linear approximation
operator $\Pi_p$, common to all tasks and actions, with
$(\Pi_p g)(x)=\phi(x)^\top b_p(g)$ such that
\[
\sup_{g\in\Lambda(\kappa,c)}
\|g-\Pi_p g\|_\infty
\le C_{\mathrm{app}}p^{-\kappa/d},
\]
where $C_{\mathrm{app}}$ is independent of $p$.
For each $k,a$, the map
$x'\mapsto b_p\!\left(\rho^{(k)}(x'\mid\cdot,a)\right)$
is measurable and componentwise integrable.

\item \textbf{Well-conditioned sieve design.}
\label{assump:covariance}
For each task $k$, define the population feature
second-moment matrix
\[
\Sigma^{(k)}
=
\frac{1}{T}\mathbb E\!\left[
\sum_{t=0}^{T-1}
\xi(X_{i,t}^{(k)},A_{i,t}^{(k)})
\xi^\top(X_{i,t}^{(k)},A_{i,t}^{(k)})
\right].
\]
There exists a constant $c_\Sigma\ge1$ such that
\[
c_\Sigma^{-1}
\le \lambda_{\min}(\Sigma^{(k)})
\le \lambda_{\max}(\Sigma^{(k)})
\le c_\Sigma,
\qquad k\in\{0\}\cup[K].
\]
The sieve features also satisfy
$\sup_{x\in\calX}\|\phi(x)\|_2^2\lesssim p$ and
$\sup_{x\in\calX,\ a\in\calA}\|\xi(x,a)\|_2^2\lesssim p$.

\end{enumerate}
\end{assumption}

For each task $k$, let $\beta_r^{(k)}\in\mathbb R^q$
and $\beta_\rho^{(k)}(x')\in\mathbb R^q$ be the
action-stacked coefficients obtained by applying $b_p$
to $r^{(k)}(\cdot,a)$ and
$\rho^{(k)}(x'\mid\cdot,a)$, respectively.
Conditional on the preceding fitted functions, define
\[
\beta_\tau^{(k)}
=
\beta_r^{(k)}
+
\gamma\int_{\calX}
\max_{a'\in\calA}
\widehat Q_{\tau-1}^{(k)}(x',a')
\,\beta_\rho^{(k)}(x')\,dx'.
\]
Then $\xi^\top(x,a)\beta_\tau^{(k)}$ is the induced
sieve approximation to
$f_\tau^{(k)}=\calT^{(k)}\widehat Q_{\tau-1}^{(k)}$.
For common transition kernels,
$\beta_\rho^{(k)}(x')=\beta_\rho(x')$ for all tasks.

For each iteration $\tau$ and task $k$, define
$\widehat\Sigma_\tau^{(k)} = \frac{1}{n_k}(Z_\tau^{(k)})^\top Z_\tau^{(k)}$.
Set $c_{\mathrm G}=(2c_\Sigma)^{-1}$ and
$C_{\mathrm G}=2c_\Sigma$, and let
\[
\mathcal E_{\Sigma,\tau}
:=
\bigcap_{k=0}^{K}
\left\{
c_{\mathrm G}I_q
\preceq \widehat\Sigma_\tau^{(k)}
\preceq C_{\mathrm G}I_q
\right\}.
\]

\begin{lemma}[Uniform empirical Gram bounds]
\label{lem:uniform-empirical-gram}
Suppose Assumption~\ref{assump:covariance} holds and
complete independent trajectories are assigned to
iterations as described above. Let
$J_{\min} = \min_{0\le k\le K}\frac{I^{(k)}}{\Upsilon}$.
For a constant $c>0$ independent of the sample sizes
and sieve dimension,
\[
\mathbb P\!\left(
\bigcup_{\tau=1}^{\Upsilon}
\mathcal E_{\Sigma,\tau}^{c}
\right)
\le
2(K+1)\Upsilon q
\exp\!\left(-c\frac{J_{\min}}{p}\right).
\]
Consequently, if $K,\Upsilon$ are fixed and
$p\log(q\vee2)/J_{\min}=o(1)$, then
$\mathbb P\!\left( \bigcap_{\tau=1}^{\Upsilon} \mathcal E_{\Sigma,\tau} \right)
\longrightarrow1$.
\end{lemma}

The common approximation rule defines the task-specific
Bellman coefficients in the same sieve coordinates.
The lower density bound in
Assumption~\ref{assump:ergodicity} gives
$\operatorname{Leb}(\calX)\le c_\mu^{-1}$.
Together with the approximation condition, this yields
a Bellman approximation error of order $p^{-\kappa/d}$
on the bounded-iterate event required in the theorems.
Lemma~\ref{lem:uniform-empirical-gram} supplies the
empirical spectral bounds needed for the joint fit
under its sample-size condition. It uses independence
across complete trajectories while allowing dependence
within each trajectory.

\begin{assumption}
\label{assump:joint-transfer}
For $\mu>0$, the following conditions hold for the target-source
correction and the profiled joint regression.
\begin{enumerate}
\renewcommand{\labelenumi}{(\roman{enumi})}
\renewcommand{\theenumi}{\theassumption(\roman{enumi})}

\item \textbf{Sparse target-source correction.}
\label{assump:l1-correction}
Let
$\overline\beta_{\tau,s} = \sum_{k=1}^K w_k\beta_\tau^{(k)}$ and
$\delta_{\tau,\Delta}^{(0)} = \beta_\tau^{(0)}-\overline\beta_{\tau,s}$,
and define
$h_\Delta = \max_{1\le\tau\le\Upsilon} \|\delta_{\tau,\Delta}^{(0)}\|_1$ and
$S_{\tau,\Delta} = \operatorname{supp}\bigl(\delta_{\tau,\Delta}^{(0)}\bigr)$.
There exists an integer $s\ge1$ such that
\[
\max_{1\le\tau\le\Upsilon}|S_{\tau,\Delta}|
=
\max_{1\le\tau\le\Upsilon}
\|\delta_{\tau,\Delta}^{(0)}\|_0
\le s.
\]

\item \textbf{Compatibility of the profiled correction.}
\label{assump:compatibility}
On the event that the source empirical Gram matrices
are nonsingular, define
$\widehat\Omega_{\tau,s} = \left[ \sum_{k=1}^K
w_k\bigl(\widehat\Sigma_\tau^{(k)}\bigr)^{-1} \right]^{-1}$ and $\widehat H_\tau =
\widehat\Sigma_\tau^{(0)} +\mu\widehat\Omega_{\tau,s}$.
Let $\kappa_\mu=\mu/(1+\mu)$ and
$M_\tau = \widehat\Sigma_\tau^{(0)} - \widehat\Sigma_\tau^{(0)} \widehat H_\tau^{-1}
\widehat\Sigma_\tau^{(0)}$.
There exists a constant $\phi_{\mathrm{prof}}>0$,
independent of the dimension, sample sizes, and iteration,
such that, with probability tending to one,
uniformly over $\tau=1,\ldots,\Upsilon$,
\[
\|u_{S_{\tau,\Delta}}\|_1^2
\le
\frac{|S_{\tau,\Delta}|}
{\kappa_\mu\phi_{\mathrm{prof}}^2}
u^\top M_\tau u
\]
for every $u\in\mathbb R^{mp}$ satisfying
$\|u_{S_{\tau,\Delta}^c}\|_1 \le 3\|u_{S_{\tau,\Delta}}\|_1$.
Here $u_S$ is the subvector indexed by $S$, and $S^c$ is its complement.
The requirement is vacuous when $S_{\tau,\Delta}=\varnothing$.

\item \textbf{Coordinatewise profile stability.}
\label{assump:profile-stability}
Define
$\mathcal A_{\tau,0} = I-\widehat\Sigma_\tau^{(0)}\widehat H_\tau^{-1}$,
$\mathcal A_{\tau,s} = \mu\widehat\Sigma_\tau^{(0)}\widehat H_\tau^{-1}$,
and, for each source task,
$\mathcal B_{\tau,k} = \widehat\Omega_{\tau,s}
\bigl(\widehat\Sigma_\tau^{(k)}\bigr)^{-1}$.
There exists a constant $C_{\mathrm{prof}}>0$,
independent of $q$, the sample sizes, and $\tau$,
such that, with probability tending to one,
\[
\max_{1\le\tau\le\Upsilon}
\max\left\{
\kappa_\mu^{-1}
\|\mathcal A_{\tau,0}\|_{\infty\to\infty},
\;
\kappa_\mu^{-1}
\|\mathcal A_{\tau,s}\|_{\infty\to\infty},
\;
\max_{k\in[K]}
\|\mathcal B_{\tau,k}\|_{\infty\to\infty}
\right\}
\le C_{\mathrm{prof}},
\]
where $\|A\|_{\infty\to\infty}=\max_i\sum_j|A_{ij}|$.

\end{enumerate}
\end{assumption}

The joint $\ell_1$ penalty acts on the target-source
correction, whose support size and magnitude are measured
by $s$ and $h_\Delta$, respectively.
Compatibility provides curvature along the corresponding
sparse cone, while coordinatewise profile stability
controls the amplification of regression score errors
when the source coefficients are profiled out.
Since $h_\Delta$ depends on preceding fitted iterates,
it may be random. Throughout, $X_n=O_p(R_n)$ for a
positive random rate $R_n$ means $X_n/R_n=O_p(1)$.

\subsection{Transition-Homogeneous Tasks}

We first consider tasks sharing a transition kernel
$P^{(k)}(\cdot\mid x,a)=P(\cdot\mid x,a)$,
while allowing task-specific reward functions.
Write $n_{\mathrm{all}}=n_0+n_s$ and
$n_{\min}=\min_{0\le k\le K}n_k$.
For the joint update, take $\mu=n_s/n_0$ and
$\lambda_\Delta=c_\Delta\sqrt{n_s\log q/(n_0n_{\mathrm{all}})}$,
where $c_\Delta>0$ is sufficiently large.
Since the profiled correction loss has curvature of order
$\kappa_\mu$, its effective regularization level is
$\overline\lambda_\Delta=\lambda_\Delta/\kappa_\mu
=c_\Delta\sqrt{n_{\mathrm{all}}\log q/(n_0n_s)}$.
The quantity $h_\Delta$ measures the largest $\ell_1$ gap,
over iterations, between the target Bellman coefficients and
the weighted source aggregate; $s$ bounds the correction's
support size. Define the corresponding correction rate
\[
\Delta_{s,h_\Delta}
=
\sqrt{s}\,\overline\lambda_\Delta
\wedge
\sqrt{h_\Delta\overline\lambda_\Delta}
\wedge h_\Delta.
\]
For a target policy $\pi$, write
$v^{\pi,(0)}=\mathbb E_{X_0\sim\nu^{(0)}}[V^{\pi,(0)}(X_0)]$
for its integrated value.

\begin{theorem}[Policy value loss under homogeneous transitions]
\label{thm:l1-joint-transfer-fqi}
Suppose Assumptions~\ref{assump:data-process}--
\ref{assump:joint-transfer} hold, all tasks share the
transition kernel $P(\cdot\mid x,a)$, and the sieve basis
consists of B-splines or wavelets satisfying the standard
basis properties in Lemma~E.1 of the supplementary material
to \citet{Shi2022ValueInference}.
Let $K,\Upsilon$ be fixed and, for some deterministic
constant $B<\infty$, assume
\[
\frac{q\{\log^2(n_{\min})+T\log(q\vee2)\}}{n_{\min}}=o(1),
\qquad
\mathbb P\!\left(
\max_{\substack{0\le k\le K\\0\le\tau\le\Upsilon}}
\|\widehat Q_\tau^{(k)}\|_\infty\le B
\right)\longrightarrow1.
\]
Let $\pi^{*,(0)}$ be an optimal target policy.
If $\max_{0\le k\le K}\|\widehat Q_0^{(k)}\|_\infty
\le R_{\max}/(1-\gamma)$, the greedy policy
$\widehat\pi_\Upsilon^{(0)}$ returned by
Algorithm~\ref{alg:joint-transfer-fqi} satisfies
\[
v^{\pi^{*,(0)},(0)}-v^{\widehat\pi_\Upsilon^{(0)},(0)}
=\frac{1}{(1-\gamma)^2}\,O_p\!\Bigl(
\underbrace{p^{-\kappa/d}}_{
\substack{\text{sieve}\\\text{bias}}}
\;+\;
\underbrace{\sqrt{q/n_{\mathrm{all}}}}_{
\substack{\text{pooled}\\\text{estimation}}}
\;+\;
\underbrace{\Delta_{s,h_\Delta}}_{
\substack{\text{target-source}\\\text{correction}}}
\;+\;
\underbrace{\gamma^\Upsilon R_{\max}}_{
\substack{\text{iteration}\\\text{error}}}
\Bigr).
\]
\end{theorem}

Theorem~\ref{thm:l1-joint-transfer-fqi} bounds the policy value
loss by three statistical terms and an iteration remainder.
The first term, $p^{-\kappa/d}$, arises from approximating the
Bellman regression functions using the sieve basis.
Increasing $p$ reduces this approximation error but also
increases the coefficient dimension $q=mp$.
The second term, $\sqrt{q/n_{\mathrm{all}}}$, reflects the
estimation accuracy supported by the combined target and
source samples. The third term, $\Delta_{s,h_\Delta}$,
accounts for correcting the discrepancy between the target
and the weighted source aggregate.
The final term, $\gamma^\Upsilon R_{\max}$, represents the
residual effect of initialization after $\Upsilon$ Bellman
updates. The common factor $(1-\gamma)^{-2}$ reflects
error propagation through these updates and its effect on
policy value.

The correction bound adapts to both the sparsity and the
magnitude of the aggregate discrepancy.
In the source-rich regime $n_s\gg n_0$, we have
$n_{\mathrm{all}}\asymp n_s$ and
$\overline\lambda_\Delta\asymp\sqrt{\log q/n_0}$, giving
\[
\Delta_{s,h_\Delta}
\asymp
\sqrt{\frac{s\log q}{n_0}}
\wedge
\sqrt{h_\Delta}
\left(\frac{\log q}{n_0}\right)^{1/4}
\wedge
h_\Delta.
\]
The first branch exploits the sparse support of the
aggregate correction; the other two use its $\ell_1$
magnitude. Specifically, in the original definition of
$\Delta_{s,h_\Delta}$, the linear branch applies when
$h_\Delta\le\overline\lambda_\Delta$, while the sparse
branch applies when
$h_\Delta\ge s\overline\lambda_\Delta$.
Between these regimes, the correction rate is the
geometric mean of the discrepancy magnitude and the
effective regularization level.

This decomposition clarifies the distinct roles of source
and target data. Once source data are abundant, the pooled
estimation term continues to decrease with the combined
sample size, whereas the regularization scale for target
correction is primarily determined by $n_0$.
For example, if additionally $s\log q=o(q)$, then
$\sqrt{q/n_{\mathrm{all}}}+\Delta_{s,h_\Delta}
=o(\sqrt{q/n_0})$.
Thus, the estimation component of the bound can be smaller
than the target-sample scale $\sqrt{q/n_0}$.
A small aggregate discrepancy provides another favorable
case through $\Delta_{s,h_\Delta}\le h_\Delta$.
The overall value bound may nevertheless be governed by
sieve approximation or the finite-iteration remainder.

The bound retains the approximation, pooled estimation,
and iteration rates of \citet{Chen2026BatchQLearning}.
Under our structural conditions, the source-rich
correction bound admits the branch
$\sqrt{s\log q/n_0}$, improving the dimension dependence
when $s\ll q$ for a fixed action space.
Moreover, $h_\Delta$ measures discrepancy relative to
the weighted source aggregate, allowing differences
across sources to offset one another.
The discrepancy-dependent branches
$\sqrt{h_\Delta}(\log q/n_0)^{1/4}\wedge h_\Delta$
also have counterparts in transfer regression
\citep{Li2022TransLasso}, reflecting a shared statistical
structure: pooled observations support estimation
across tasks, while target observations control the
precision of discrepancy correction.

\subsection{Transition-Heterogeneous Tasks}

In the transition-heterogeneous setting, we consider tasks with transition kernels
$P^{(k)}(\cdot\mid x,a)$ and corresponding densities
$\rho^{(k)}(\cdot\mid x,a)$ that may differ across tasks.
Reward functions $r^{(k)}$ may also vary, so task
heterogeneity arises from both immediate rewards
and transition dynamics.
Using the common sieve representations, define the
discrepancy between source task $k$ and the target by
\[
h_k
=
\left\|\beta_r^{(k)}-\beta_r^{(0)}\right\|_1
+
\int_{\mathcal X}
\left\|\beta_\rho^{(k)}(x')-\beta_\rho^{(0)}(x')\right\|_1
\,dx'.
\]
The weighted discrepancy
$h_{\mathrm w}=\sum_{k=1}^K w_k h_k$
summarizes heterogeneity in the underlying rewards and
transitions. The earlier quantity $h_\Delta$ measures
the largest aggregate Bellman-coefficient discrepancy
over fitted iterations.

To control how differences between fitted value functions
enter the Bellman coefficients, define the target
transition coefficient operator
$\mathcal K_\rho^{(0)}g
=\int_{\mathcal X}g(x')\beta_\rho^{(0)}(x')\,dx'$
and impose the stability condition
\begin{equation}
\left\|\mathcal K_\rho^{(0)}g\right\|_1
\le C_\rho\|g\|_{L_1},
\qquad g\in L_1(\mathcal X),
\label{eq:transition-coefficient-stability}
\end{equation}
where $C_\rho<\infty$ does not depend on the sieve dimension.
Retain the choices of $\mu$, $\lambda_\Delta$, and
$\overline\lambda_\Delta$ from the homogeneous setting,
and define the correction rate
$\Delta_{s,h_{\mathrm w}}
=\sqrt{s}\,\overline\lambda_\Delta
\wedge\sqrt{h_{\mathrm w}\overline\lambda_\Delta}
\wedge h_{\mathrm w}$.

\begin{theorem}[Policy value loss under heterogeneous transitions]
\label{thm:heterogeneous-l1-joint-transfer-fqi}
Suppose Assumptions~\ref{assump:data-process}--
\ref{assump:joint-transfer} and the transition-coefficient
stability condition~\eqref{eq:transition-coefficient-stability}
hold. Suppose the sieve basis consists of B-splines or
wavelets satisfying the standard basis properties in
Lemma~E.1 of the supplementary material to
\citet{Shi2022ValueInference}.
Let $K,\Upsilon$ be fixed and $n_s\gtrsim n_0$.
For some deterministic constant $B<\infty$, assume
\[
\frac{q\{\log^2(n_{\min})+T\log(q\vee2)\}}{n_{\min}}=o(1),
\qquad
\mathbb P\!\left(
\max_{\substack{0\le k\le K\\0\le\tau\le\Upsilon}}
\|\widehat Q_\tau^{(k)}\|_\infty\le B
\right)\longrightarrow1.
\]
Suppose all tasks use a common initialization
$\widehat\beta_0^{(k)}=\widehat\beta_0^{(0)}$ for $k\in[K]$,
with
$\|\widehat Q_0^{(0)}\|_\infty\le R_{\max}/(1-\gamma)$.
Let $\pi^{*,(0)}$ be an optimal target policy.
Then the greedy target policy $\widehat\pi_\Upsilon^{(0)}$
returned by Algorithm~\ref{alg:joint-transfer-fqi} satisfies
\begin{equation}
v^{\pi^{*,(0)},(0)}-v^{\widehat\pi_\Upsilon^{(0)},(0)}
=
\frac{1}{(1-\gamma)^2}\,O_p\!\Bigl(
p^{-\kappa/d}
+\sqrt{q/n_{\mathrm{all}}}
+\Delta_{s,h_{\mathrm w}}
+\gamma^\Upsilon R_{\max}
\Bigr).
\end{equation}
\end{theorem}

Compared with Theorem~\ref{thm:l1-joint-transfer-fqi},
Theorem~\ref{thm:heterogeneous-l1-joint-transfer-fqi}
retains the same approximation, pooled estimation,
and iteration rates.
The correction term is now characterized by
$h_{\mathrm w}$, which summarizes weighted differences in
rewards and transition dynamics between the source tasks
and the target.
The additional stability and common-initialization
conditions allow these primitive differences to be
controlled through successive Bellman regressions.

Under $n_s\gtrsim n_0$, we have
$n_{\mathrm{all}}\asymp n_s$ and
$\overline\lambda_\Delta\asymp\sqrt{\log q/n_0}$.
A sufficient condition for the correction term to be
no larger in order than the pooled estimation term is
\[
n_0\gtrsim
\frac{h_{\mathrm w}^{\,2}n_s^2\log q}{q^2}.
\]
Indeed, the middle branch then gives
$\Delta_{s,h_{\mathrm w}}
\le\sqrt{h_{\mathrm w}\overline\lambda_\Delta}
\lesssim\sqrt{q/n_{\mathrm{all}}}$,
so the statistical part of the value bound is governed
by sieve approximation and pooled estimation.
For fixed sieve dimension and source sample size,
this sufficient condition becomes more demanding as
$h_{\mathrm w}$ increases, highlighting the role of
target observations in correcting task differences.
The sparse branch $\sqrt{s}\,\overline\lambda_\Delta$
provides an alternative bound through the support size
of the aggregate Bellman correction.

\section{Empirical Studies}
\label{sec:empirical}
\subsection{Monte Carlo Simulations}
\label{sec:mc}

We conduct simulation studies to evaluate the effectiveness
of DFQI (Algorithm~\ref{alg:joint-transfer-fqi}) for offline
transfer learning across related tasks. Following the simulation designs of
\citet{Shi2022ValueInference,Chen2026BatchQLearning}, we consider the state
space $\mathcal{X}=[-1,1]^3$ and action space $\mathcal{A}=\{-1,+1\}$.
Starting from $X_0=\tanh(Z_0)$, $Z_0\sim N(0,I_3)$, we sample actions
uniformly from $\mathcal{A}$ and generate
$X_{t+1}=\tanh\{0.75\operatorname{diag}(A_t,-A_t,A_t)X_t+\xi_t\mathbf{1}_3\}$,
with reward $R_t=A_tX_t^\top M X_t+\varepsilon_{r,t}$, where $M$ is the
task-specific reward matrix and $\xi_t$ and $\varepsilon_{r,t}$ are
independent $N(0,0.5^2)$ noises.

Using this setting, we construct one target task ($k=0$) and one source
task ($k=1$). The two tasks share the state transition mechanism, with
task heterogeneity introduced through their reward matrices. The entries
of the target matrix $M^{(0)}$ are drawn independently, with
$M_{ii}^{(0)}\sim N(0,1)$ and $M_{ij}^{(0)}\sim N(0,0.25^2)$ for $i\ne j$.
For the source task, we set $M^{(1)}=M^{(0)}+\Delta$, where
$\Delta_{ij}\overset{\mathrm{iid}}{\sim}N(0,\sigma^2)$.
We vary $\sigma\in\{0.25,0.5,0.75,1\}$ to control task discrepancy and
consider $I^{(0)}\in\{20,40,60,80\}$ target trajectories and
$I^{(1)}\in\{10,20,40,80\}$ source trajectories, each of length five.

We compare three methods in our experiment: (i) No transfer,
which applies FQI to the target data alone; (ii) batch
$Q$-learning, the two-step method of
\citet{Chen2026BatchQLearning}, which pools source and
target data and then estimates a target-specific correction
at each fitted $Q$-iteration; and (iii) Direct Transfer Fitted $Q$-Iteration, our proposed
DFQI method (Algorithm~\ref{alg:joint-transfer-fqi}).
All methods use discount factor $\gamma=0.25$ and an additive
quadratic B-spline sieve with six basis functions per state
coordinate and separate coefficients for each action.
Initial coefficients are drawn independently from
$N(0,0.01^2)$. In each replication, we estimate
$\mathbb{E}|\widehat Q(X,A)-Q^*(X,A)|$ by averaging over
200 test state--action pairs drawn independently from the
initial state--action distribution, with $Q^*$ approximated
by Monte Carlo simulation.

\begin{figure}[p]
  \centering
  \includegraphics[width=\textwidth]{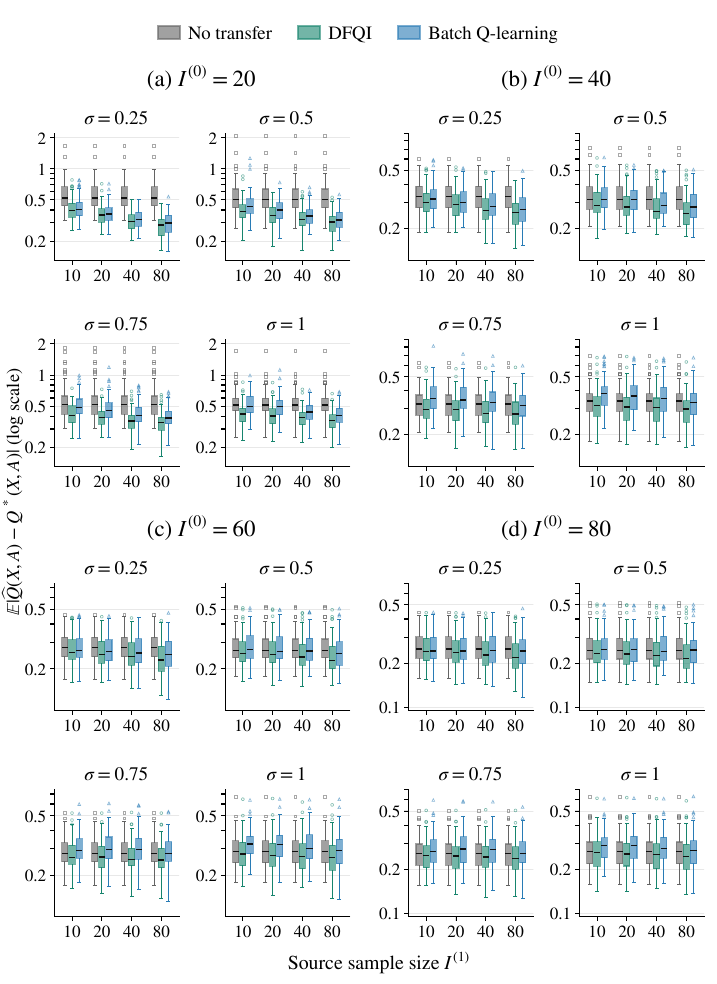}
  \caption{Boxplots of the estimation errors
  $\mathbb{E}|\widehat Q(X,A)-Q^*(X,A)|$ with
  target sample sizes $I^{(0)}=20,40,60,80$ in blocks (a)--(d), respectively,
  and source sample sizes $I^{(1)}\in\{10,20,40,80\}$.
  The parameter $\sigma$ above each panel controls task discrepancy
  $\Delta=M^{(1)}-M^{(0)}$. Each boxplot summarizes 100 replications.
  Vertical axes use logarithmic scales with panel-specific limits.}
  \label{fig:mc-all}
\end{figure}

Figure~\ref{fig:mc-all} shows that DFQI achieves lower median estimation error than
both two-step batch $Q$-learning and No transfer across all target and
source sample sizes and task discrepancies considered. Its interquartile
ranges are also generally shifted toward lower errors. These results
support the effectiveness of DFQI in learning from source data under
different degrees of task heterogeneity.

Increasing the target sample size reduces the median estimation error of all three
methods, with diminishing gains. The reduction is largest when
$I^{(0)}$ increases from 20 to 40, smaller from 40 to 60, and more modest
from 60 to 80. These equal increments of 20 trajectories correspond to
progressively smaller proportional increases in target data, from
$100\%$ to $50\%$ and approximately $33\%$, respectively.


\subsection{Real Data Analysis}
\label{sec:real-data}

We evaluate DFQI in a semisynthetic study based on adult sepsis data
from MIMIC-III \citep{Johnson2016MIMICIII}. We use the preprocessed
cohort selected following \citet{Komorowski2018AIClinician} and adopt
the evaluation framework of \citet{Chen2026BatchQLearning}.

States are constructed from 44 demographic and clinical variables in
the preceding four-hour window and reduced to 10 principal components
after transformation and normalization. Intravenous fluid volumes and
vasopressor doses are each discretized into three levels, giving nine
actions. Rewards combine penalties for deviations of 15 physiological
measurements from prespecified ranges with a mortality-based terminal
component \citep{Prasad2017VentilationWeaning,Komorowski2018AIClinician}.
At the end of each observed record, an additional reward of $-50$ is
assigned for recorded 90-day mortality and $+50$ otherwise.

We retain only transitions between consecutive four-hour windows,
yielding 218,298 transitions from 18,369 episodes. We designate adults
aged 65 years or older as the target cohort and those aged 18--49 years
as the source cohort, comprising 9,558 and 3,475 episodes, respectively.
For each cohort, we calibrate an environment by estimating its reward
function, transition distribution, termination probability, and
initial-state distribution.

From each calibrated environment, we generate a fixed pool of
2,000 training trajectories under a uniform random behavior policy.
In each replication, we sample
$I^{(0)}\in\{100,500\}$ target trajectories and
$I^{(1)}\in\{250,500,\ldots,2000\}$ source trajectories
without replacement. We compare the same three methods as in
the Monte Carlo study, using quadratic B-spline bases with
separate coefficients for each action and discount factor
$\gamma=0.6$.

Generated trajectories contain at most four decisions, with earlier
termination allowed. Each learned target policy $\widehat\pi$ is
evaluated by its mean discounted return over 200 newly generated
target trajectories. We report policy regret
$v^{\pi^*}-v^{\widehat\pi}$, where $\pi^*$ is an optimal target
policy and $v^\pi$ denotes the expected discounted return of policy
$\pi$ in the calibrated target environment. Results are averaged
across replications; lower regret indicates better performance.

\begin{figure}[tbp]
  \centering
  \includegraphics[width=\textwidth]{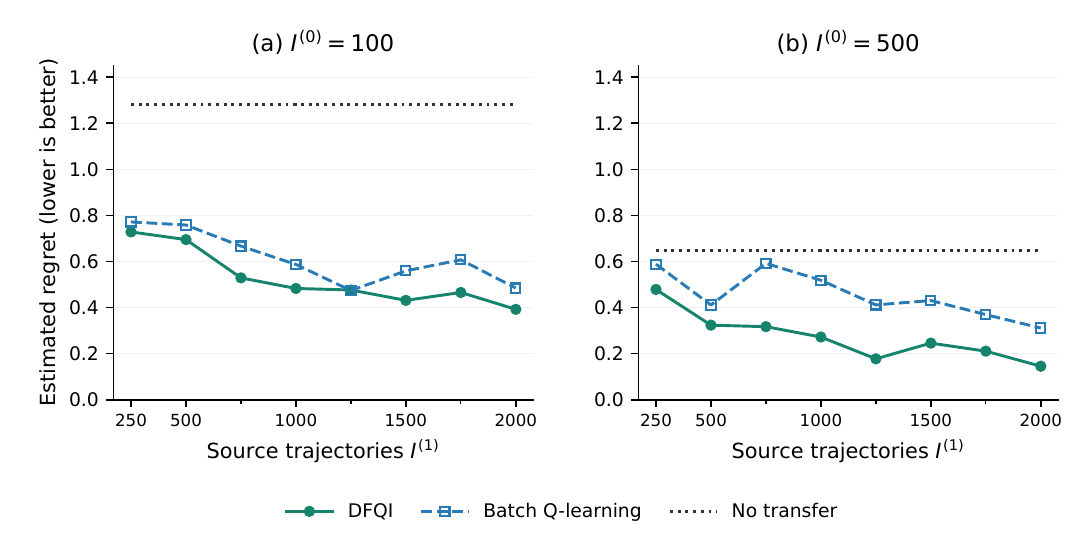}
    \caption{Estimated target-policy regret
    $v^{\pi^*}-v^{\widehat\pi}$ against the number of source
    trajectories $I^{(1)}$ in the calibrated environment,
    with $I^{(0)}=100$ (left) and $I^{(0)}=500$ (right).
    Curves show means across replications; lower values are better.}
  \label{fig:real-data}
\end{figure}

Figure~\ref{fig:real-data} presents the mean estimated policy
regret $v^{\pi^*}-v^{\widehat\pi}$ of the three methods across
target and source sample-size configurations. With
$I^{(0)}=100$ target trajectories, our proposed DFQI
achieves lower regret than target-only FQI throughout the
source-sample range and generally improves on batch
$Q$-learning, with the only exception at
$I^{(1)}=1250$ where their means are
nearly equal. With $I^{(0)}=500$ target trajectories,
DFQI has the lowest mean regret at every source sample
size, with a clearer separation from batch $Q$-learning.
For both target sample sizes, the regret of the two transfer
methods declines overall as more source trajectories are used,
despite some nonmonotonicity. These findings illustrate the
benefit of joint transfer for target-policy learning in the
calibrated environment.

\section{Conclusion}
\label{sec:conclusion}

This paper develops Direct Transfer Fitted $Q$-Iteration
(DFQI) for offline reinforcement learning. At each iteration,
DFQI integrates target and source $Q$-function estimation
within a single optimization problem, using an $\ell_1$
fusion penalty to share information across tasks while
accommodating their heterogeneity. 
We establish estimation and policy regret bounds under
sieve approximation for tasks with shared or heterogeneous transition
kernels, clarifying the roles of sample size and task discrepancy.
Monte Carlo simulations and a semisynthetic study calibrated to
MIMIC-III illustrate the benefits of the proposed framework for
$Q$-function estimation and policy learning.

\bibliography{refs-DFQI}

\clearpage

\appendix
\numberwithin{equation}{section}
\section{Proof for Theoretical Results}
\label{app:proofs}

\subsection{Proof of Lemma~\ref{lem:q-discrepancy-decomposition}}
\begin{proof}[Proof of Lemma~\ref{lem:q-discrepancy-decomposition}]
By the Bellman optimality equation, for the source task $k$ we have
\[
Q^{*(k)}(x,a)
=
r^{(k)}(x,a)
+
\gamma
\int_{\calX}
\max_{a'\in\calA}
Q^{*(k)}(x',a')
\rho^{(k)}(x'\mid x,a)\,dx' .
\]
Similarly, for the target task,
\[
Q^{*(0)}(x,a)
=
r^{(0)}(x,a)
+
\gamma
\int_{\calX}
\max_{a'\in\calA}
Q^{*(0)}(x',a')
\rho^{(0)}(x'\mid x,a)\,dx' .
\]
Subtracting the target Bellman equation from the source Bellman equation yields
\[
Q^{*(k)}(x,a)-Q^{*(0)}(x,a)
=
r^{(k)}(x,a)-r^{(0)}(x,a)
\]
\[
+
\gamma
\int_{\calX}
\left[
\max_{a'\in\calA}
Q^{*(k)}(x',a')
\rho^{(k)}(x'\mid x,a)
-
\max_{a'\in\calA}
Q^{*(0)}(x',a')
\rho^{(0)}(x'\mid x,a)
\right]dx' .
\]
Adding and subtracting
\[
\gamma
\int_{\calX}
\max_{a'\in\calA}
Q^{*(k)}(x',a')
\rho^{(0)}(x'\mid x,a)\,dx'
\]
inside the last display gives
\[
Q^{*(k)}(x,a)-Q^{*(0)}(x,a)
=
\delta_r^{(k)}(x,a)
+
\gamma
\int_{\calX}
\max_{a'\in\calA}
Q^{*(k)}(x',a')
\left[
\rho^{(k)}(x'\mid x,a)
-
\rho^{(0)}(x'\mid x,a)
\right]dx'
\]
\[
\qquad\qquad
+
\gamma
\int_{\calX}
\left[
\max_{a'\in\calA}
Q^{*(k)}(x',a')
-
\max_{a'\in\calA}
Q^{*(0)}(x',a')
\right]
\rho^{(0)}(x'\mid x,a)\,dx' .
\]
Using the definitions
\[
\delta_Q^{(k)}(x,a)
=
Q^{*(k)}(x,a)-Q^{*(0)}(x,a),
\]
\[
\delta_r^{(k)}(x,a)
=
r^{(k)}(x,a)-r^{(0)}(x,a),
\]
and
\[
\delta_\rho^{(k)}(x'\mid x,a)
=
\rho^{(k)}(x'\mid x,a)
-
\rho^{(0)}(x'\mid x,a),
\]
we obtain the desired decomposition.
\end{proof}

\subsection{Proof of Proposition~\ref{prop:convexity-solvability}}
\label{app:proof-convexity-solvability}

\begin{proof}
Fix an iteration \(\tau\). Conditional on the previous fitted iterates, the
pseudo-responses \(\{\widehat y_\tau^{(k)}\}_{k=0}^K\) are fixed. Let
\(q=mp\), and define the stacked parameter
\[
\theta
=
\left(
\beta^{(0)\top},
\beta^{(1)\top},
\ldots,
\beta^{(K)\top}
\right)^\top
\in
\mathbb R^{(K+1)q}.
\]
Define
\[
\widetilde Z_\tau
=
\operatorname{diag}
\left(
n_0^{-1/2}Z_\tau^{(0)},
\sqrt{\frac{\mu w_1}{n_1}}Z_\tau^{(1)},
\ldots,
\sqrt{\frac{\mu w_K}{n_K}}Z_\tau^{(K)}
\right),
\]
and
\[
\widetilde y_\tau
=
\left(
n_0^{-1/2}\widehat y_\tau^{(0)\top},
\sqrt{\frac{\mu w_1}{n_1}}\widehat y_\tau^{(1)\top},
\ldots,
\sqrt{\frac{\mu w_K}{n_K}}\widehat y_\tau^{(K)\top}
\right)^\top .
\]
Moreover, define the contrast matrix
\[
D
=
\left[
I_q,\,
-w_1I_q,\,
-w_2I_q,\,
\ldots,\,
-w_KI_q
\right].
\]
Then
\[
D\theta
=
\beta^{(0)}
-
\sum_{k=1}^K
w_k\beta^{(k)}.
\]
Therefore the sieve joint update can be written as
\[
\widehat\theta_\tau
\in
\arg\min_{\theta\in\mathbb R^{(K+1)q}}
\left\{
\frac12
\left\|
\widetilde y_\tau-\widetilde Z_\tau\theta
\right\|_2^2
+
\lambda_\Delta
\left\|
D\theta
\right\|_1
\right\}.
\]

The first term is a convex quadratic function of \(\theta\). The second term is
also convex, since it is the composition of the convex norm \(\|\cdot\|_1\)
with the linear map \(D\). Hence the full objective is convex.

It remains to show existence of a minimizer. If \(\lambda_\Delta=0\), the
problem reduces to ordinary least squares with design matrix
\(\widetilde Z_\tau\). Since the column space of \(\widetilde Z_\tau\) is a
closed subspace of a finite-dimensional Euclidean space, the projection of
\(\widetilde y_\tau\) onto this column space exists. Hence the least-squares
problem has at least one minimizer.

Now suppose \(\lambda_\Delta>0\). Consider the linear map
\[
A:
\mathbb R^{(K+1)q}
\to
\mathbb R^{n_0+n_s+q},
\qquad
A\theta
=
\left(
\widetilde Z_\tau\theta,
D\theta
\right).
\]
Let
\[
\mathcal M
=
A\mathbb R^{(K+1)q}
=
\left\{
\left(
\widetilde Z_\tau\theta,
D\theta
\right):
\theta\in\mathbb R^{(K+1)q}
\right\}.
\]
Because \(\mathcal M\) is a finite-dimensional linear subspace, it is closed.
The original problem is equivalent to minimizing
\[
\Psi(u,v)
=
\frac12
\left\|
\widetilde y_\tau-u
\right\|_2^2
+
\lambda_\Delta
\|v\|_1
\]
over \((u,v)\in\mathcal M\).

Let \(\{(u_\ell,v_\ell)\}_{\ell\ge1}\subset\mathcal M\) be a minimizing
sequence. Since
\[
\Psi(u_\ell,v_\ell)
\le
\Psi(0,0)+o(1),
\]
we have
\[
\frac12
\left\|
\widetilde y_\tau-u_\ell
\right\|_2^2
\le
\Psi(0,0)+o(1),
\]
and
\[
\lambda_\Delta
\|v_\ell\|_1
\le
\Psi(0,0)+o(1).
\]
Thus \(\{u_\ell\}\) and \(\{v_\ell\}\) are bounded. Since the ambient space is
finite-dimensional, there exists a convergent subsequence, still denoted by
\(\{(u_\ell,v_\ell)\}\), such that
\[
(u_\ell,v_\ell)\to(u_*,v_*).
\]
Because \(\mathcal M\) is closed, \((u_*,v_*)\in\mathcal M\). By continuity of
\(\Psi\),
\[
\Psi(u_*,v_*)
=
\inf_{(u,v)\in\mathcal M}
\Psi(u,v).
\]
Since \((u_*,v_*)\in\mathcal M\), there exists at least one
\(\widehat\theta_\tau\in\mathbb R^{(K+1)q}\) such that
\[
A\widehat\theta_\tau=(u_*,v_*).
\]
Therefore \(\widehat\theta_\tau\) is a minimizer of the original problem.

Finally, since the objective is convex, any minimizer satisfies the subgradient
optimality condition
\[
0
\in
\widetilde Z_\tau^\top
\left(
\widetilde Z_\tau\widehat\theta_\tau
-
\widetilde y_\tau
\right)
+
\lambda_\Delta
D^\top
\partial
\left\|
D\widehat\theta_\tau
\right\|_1.
\]
Equivalently, there exists
\[
\widehat z_\tau
\in
\partial
\left\|
D\widehat\theta_\tau
\right\|_1
\]
such that
\[
\widetilde Z_\tau^\top
\left(
\widetilde Z_\tau\widehat\theta_\tau
-
\widetilde y_\tau
\right)
+
\lambda_\Delta
D^\top
\widehat z_\tau
=
0.
\]
Writing
\[
\widehat\Delta_\tau
=
D\widehat\theta_\tau
=
\widehat\beta_\tau^{(0)}
-
\sum_{k=1}^K
w_k\widehat\beta_\tau^{(k)},
\]
the subgradient satisfies
\[
\widehat z_{\tau,j}
=
\operatorname{sign}
\left(
\widehat\Delta_{\tau,j}
\right)
\quad
\text{if }
\widehat\Delta_{\tau,j}\ne0,
\qquad
\widehat z_{\tau,j}\in[-1,1]
\quad
\text{if }
\widehat\Delta_{\tau,j}=0.
\]

Expanding the stacked KKT condition block by block gives
\[
\frac{1}{n_0}
\left(Z_\tau^{(0)}\right)^\top
\left(
Z_\tau^{(0)}
\widehat\beta_\tau^{(0)}
-
\widehat y_\tau^{(0)}
\right)
+
\lambda_\Delta
\widehat z_\tau
=
0,
\]
and, for every \(k=1,\ldots,K\),
\[
\mu w_k
\frac{1}{n_k}
\left(Z_\tau^{(k)}\right)^\top
\left(
Z_\tau^{(k)}
\widehat\beta_\tau^{(k)}
-
\widehat y_\tau^{(k)}
\right)
-
\lambda_\Delta w_k
\widehat z_\tau
=
0.
\]
This completes the proof.
\end{proof}

\subsection{Proof of Lemma~\ref{lem:uniform-empirical-gram}}

This result supplies the empirical spectral bounds used below in the
stochastic score analysis and the profiled curvature bound.

\begin{proof}[Proof of Lemma~\ref{lem:uniform-empirical-gram}]
Fix \(\tau\) and \(k\), and let \(\mathcal I_\tau^{(k)}\) be the indices
of the trajectories assigned to this iteration. Write
\(J_k=|\mathcal I_\tau^{(k)}|=I^{(k)}/\Upsilon\) and
\[
W_i^{(k)}
=
\frac1T\sum_{t=0}^{T-1}
\xi(X_{i,t}^{(k)},A_{i,t}^{(k)})
\xi(X_{i,t}^{(k)},A_{i,t}^{(k)})^\top.
\]
Then
\[
\widehat\Sigma_\tau^{(k)}
=
\frac1{J_k}\sum_{i\in\mathcal I_\tau^{(k)}}W_i^{(k)},
\qquad
\mathbb E W_i^{(k)}=\Sigma^{(k)}.
\]
The matrices \(W_i^{(k)}\) are independent across trajectories. The
sieve-basis bound gives
\(0\preceq W_i^{(k)}\preceq C_\xi pI_q\) for a constant \(C_\xi\).
Together with the population eigenvalue bounds in
Assumption~\ref{assump:covariance}, the matrix Chernoff inequality yields
\[
\mathbb P\!\left(
\lambda_{\min}(\widehat\Sigma_\tau^{(k)})<c_{\mathrm G}
\right)
+
\mathbb P\!\left(
\lambda_{\max}(\widehat\Sigma_\tau^{(k)})>C_{\mathrm G}
\right)
\le
2q\exp\!\left(-c\frac{J_k}{p}\right).
\]
A union bound over \(k\) and \(\tau\) proves the first claim. The stated
growth condition makes its right-hand side tend to zero.
\end{proof}

\subsection{Proof of Main Theorem}

\paragraph{Homogeneous tasks: conditional regression.}

We analyze the homogeneous update conditional on the fitted history,
using the notation and tuning choices of Section~5.2. The argument proceeds
from a one-step joint regression bound to the policy value loss in
Theorem~\ref{thm:l1-joint-transfer-fqi}.

Let $\mathcal F_{\tau-1}$ be the sigma-field generated by the initial fitted
functions and the data blocks $\mathcal D_1,\ldots,\mathcal D_{\tau-1}$.
Complete trajectories are assigned to iterations as in
Assumption~\ref{assump:data-process}, so the current block
$\mathcal D_\tau$ is independent of $\mathcal F_{\tau-1}$.
Write $Y_{i,t}^{(k),\tau}=\widehat Y_{i,t}^{(k),\tau}$ for the
pseudo-responses in \eqref{eq:bellman-response}, and define
\[
f_\tau^{(k)}=\mathcal T^{(k)}\widehat Q_{\tau-1}^{(k)},
\qquad a_p=p^{-\kappa/d}.
\]
The reward-innovation and Markov conditions give
\[
\mathbb E\!\left[Y_{i,t}^{(k),\tau}\mid
\mathcal F_{\tau-1},X_{i,t}^{(k)},A_{i,t}^{(k)}\right]
=f_\tau^{(k)}(X_{i,t}^{(k)},A_{i,t}^{(k)}).
\]
Thus $Y_{i,t}^{(k),\tau}=f_\tau^{(k)}(X_{i,t}^{(k)},A_{i,t}^{(k)})
+e_{i,t}^{(k),\tau}$, where the innovation is conditionally mean zero.
Its conditional sub-Gaussian bound on the bounded-history event is verified
in the proof of Lemma~\ref{lem:profiled-stochastic-score} below.

The common approximation rule defines
\[
\beta_\tau^{(k)}
=\beta_r^{(k)}+\gamma\int_{\mathcal X}
\max_{a'\in\mathcal A}\widehat Q_{\tau-1}^{(k)}(x',a')
\,\beta_\rho(x')\,dx'.
\]
These are the coefficients of the current Bellman regression approximants,
conditional on the fitted history. Let $Z_\tau^{(k)}$ have rows
$\xi^\top(X_{i,t}^{(k)},A_{i,t}^{(k)})$, and let $y_\tau^{(k)}$ collect
the current pseudo-responses. Then
\begin{equation}
y_\tau^{(k)}=Z_\tau^{(k)}\beta_\tau^{(k)}+r_\tau^{(k)}+e_\tau^{(k)},
\qquad \|r_\tau^{(k)}\|_\infty\lesssim a_p,
\label{eq:hom-regression-representation}
\end{equation}
on the bounded-history event. The sieve-remainder calculation is given
immediately below; $e_\tau^{(k)}$ denotes the innovation
and $r_\tau^{(k)}$ the approximation remainder throughout.

With $w_k=n_k/n_s$, define the population and fitted aggregate corrections by
\[
\delta_{\tau,\Delta}^{(0)}
=\beta_\tau^{(0)}-\sum_{k=1}^K w_k\beta_\tau^{(k)},
\qquad
\widehat\delta_\tau
=\widehat\beta_\tau^{(0)}-\sum_{k=1}^K w_k\widehat\beta_\tau^{(k)}.
\]
Assumption~\ref{assump:l1-correction} bounds the population correction's
support by $s$ and its $\ell_1$ norm by $h_\Delta$, uniformly in the fixed
number of iterations. We first verify the sieve remainder and derive the
profiled quadratic loss, with its projection form serving as an equivalent
verification. Using the empirical Gram bounds proved above, we next
control the stochastic scores and establish the one-step coefficient
error bound. Finally, we convert this bound into Bellman residual
control, propagate the residuals through FQI, and bound the greedy
policy's value loss.

\medskip
\noindent\textbf{Bellman sieve approximation.}\par
We verify the remainder bound in \eqref{eq:hom-regression-representation}
using the common approximation rule.
By the approximation property in Assumption~\ref{assump:smoothness}, for
any function \(g(x,a)\) satisfying \(g(\cdot,a)\in\Lambda(\kappa,c)\)
for every \(a\in\mathcal A\), let \(\beta_g\in\mathbb R^{mp}\)
stack the coefficients \(b_p(g(\cdot,a))\). Then
\[
\sup_{x\in\mathcal X,\ a\in\mathcal A}
\left|
g(x,a)-\xi^\top(x,a)\beta_g
\right|
\le
C p^{-\kappa/d}
\]
for some constant \(C>0\).

The coefficients \(\{\beta_r^{(k)}\}_{k=0}^K\) and
\(\{\beta_\rho(x')\}_{x'\in\mathcal X}\) defined above therefore satisfy
\[
\sup_{x,a,k}
\left|
r^{(k)}(x,a)
-
\xi^\top(x,a)\beta_r^{(k)}
\right|
\lesssim
p^{-\kappa/d},
\]
and, under the transition-homogeneous setting,
\[
\sup_{x,a,x'}
\left|
\rho(x'\mid x,a)
-
\xi^\top(x,a)\beta_\rho(x')
\right|
\lesssim
p^{-\kappa/d}.
\]
By definition of \(f_\tau^{(k)}=\mathcal T^{(k)}\widehat Q_{\tau-1}^{(k)}\),
we have
\[
f_\tau^{(k)}(x,a)
=
r^{(k)}(x,a)
+
\gamma
\int_{\mathcal X}
\left[
\max_{a'\in\mathcal A}
\widehat Q_{\tau-1}^{(k)}(x',a')
\right]
\rho(x'\mid x,a)\,dx'.
\]
On the bounded-history event, the preceding iterates have sup norm at most $B$.
The approximation bounds for $r^{(k)}$ and $\rho$ therefore imply
\[
\sup_{x,a,k}
\left|
f_\tau^{(k)}(x,a)
-
\xi^\top(x,a)\beta_\tau^{(k)}
\right|
\lesssim
p^{-\kappa/d}.
\]
Thus, \(\beta_\tau^{(k)}\) is the population sieve coefficient that
approximates the one-step Bellman target
\(\mathcal T^{(k)}\widehat Q_{\tau-1}^{(k)}\).

\paragraph{Direct profiling of the joint regression.}
To analyze the penalized correction, we first eliminate the source
coefficients from the joint quadratic loss.
Fix an iteration $\tau$ and condition on the preceding fitted functions.
Write $y_\tau^{(k)}$ for the held-fixed Bellman pseudo-response vector
$\widehat y_\tau^{(k)}$ in the joint update, and use the regression
representation
\[
 y_\tau^{(k)}
 =Z_\tau^{(k)}\beta_\tau^{(k)}+r_\tau^{(k)}+e_\tau^{(k)},
 \qquad k=0,\ldots,K.
\]
For candidate coefficients, define
\[
 L_{\tau,k}(\beta)
 =\frac{1}{2n_k}\|y_\tau^{(k)}-Z_\tau^{(k)}\beta\|_2^2,
 \qquad
 \alpha=(\beta^{(1)\top},\ldots,\beta^{(K)\top})^\top,
 \qquad B\alpha=\sum_{k=1}^K w_k\beta^{(k)}.
\]
The change of variables $\delta=\beta^{(0)}-B\alpha$ is one-to-one,
with inverse $\beta^{(0)}=B\alpha+\delta$. Hence the original joint
update is the minimization over $(\delta,\alpha)$ of
\[
 \ell_\tau(\delta,\alpha)+\lambda_\Delta\|\delta\|_1,
 \qquad
 \ell_\tau(\delta,\alpha)
 =L_{\tau,0}(B\alpha+\delta)
   +\mu\sum_{k=1}^K w_kL_{\tau,k}(\beta^{(k)}).
\]
For fixed $\delta$, the penalty does not depend on $\alpha$. Define
\[
 \widehat\alpha_\tau(\delta)
 \in\arg\min_\alpha\ell_\tau(\delta,\alpha),
 \qquad
 \mathcal L_{\mathrm{prof},\tau}(\delta)
 =\inf_\alpha\ell_\tau(\delta,\alpha).
\]
The infimum is attained, since this is a finite-dimensional weighted
least-squares problem. The correction component of any joint minimizer
therefore satisfies
\begin{equation}
 \widehat\delta_\tau
 \in\arg\min_{\delta\in\mathbb R^q}
 \{\mathcal L_{\mathrm{prof},\tau}(\delta)
                         +\lambda_\Delta\|\delta\|_1\}.
 \label{eq:hom-profile-minimizer}
\end{equation}
Conversely, any minimizer in \eqref{eq:hom-profile-minimizer}, together
with a corresponding $\widehat\alpha_\tau(\widehat\delta_\tau)$,
reconstructs a joint minimizer: its source coefficients are the blocks of
$\widehat\alpha_\tau(\widehat\delta_\tau)$ and its target coefficient is
$B\widehat\alpha_\tau(\widehat\delta_\tau)+\widehat\delta_\tau$.

We now obtain the coefficient form of this profiled loss on
$\mathcal E_{\Sigma,\tau}$. Put
\[
 A_{\tau,k}=\widehat\Sigma_\tau^{(k)},\qquad
 \widehat\Omega_{\tau,s}
 =\left(\sum_{k=1}^K w_kA_{\tau,k}^{-1}\right)^{-1},\qquad
 \widehat H_\tau=A_{\tau,0}+\mu\widehat\Omega_{\tau,s}.
\]
Define the raw stochastic and approximation scores, and their source
aggregates, by
\begin{equation}
 \begin{aligned}
 g_{\tau,e}^{(k)}
 &=\frac1{n_k}(Z_\tau^{(k)})^\top e_\tau^{(k)},
 &g_{\tau,r}^{(k)}
 &=\frac1{n_k}(Z_\tau^{(k)})^\top r_\tau^{(k)},
 \qquad k=0,\ldots,K,\\
 s_{\tau,e}
 &=\widehat\Omega_{\tau,s}
       \sum_{k=1}^K w_kA_{\tau,k}^{-1}g_{\tau,e}^{(k)},
 &s_{\tau,r}
 &=\widehat\Omega_{\tau,s}
       \sum_{k=1}^K w_kA_{\tau,k}^{-1}g_{\tau,r}^{(k)}.
 \end{aligned}
 \label{eq:hom-profile-scores}
\end{equation}
Write $g_\tau^{(k)}=g_{\tau,e}^{(k)}+g_{\tau,r}^{(k)}$ and
$s_\tau=s_{\tau,e}+s_{\tau,r}$.
Let $\alpha_\tau^*$ collect the source regression coefficients
$\beta_\tau^{(1)},\ldots,\beta_\tau^{(K)}$, and write
\[
 \delta_\tau^*=\delta_{\tau,\Delta}^{(0)}
 =\beta_\tau^{(0)}-\sum_{k=1}^K w_k\beta_\tau^{(k)}.
\]
For source perturbations $u^{(k)}=\beta^{(k)}-\beta_\tau^{(k)}$,
let $v=\sum_kw_ku^{(k)}$ and $h=\delta-\delta_\tau^*$.
The target perturbation is then $v+h$.

For a fixed aggregate perturbation $v$, profiling the source quadratic
loss subject to $\sum_kw_ku^{(k)}=v$ gives
\begin{equation}
 \begin{aligned}
 &\min_{\sum_kw_ku^{(k)}=v}
 \mu\sum_{k=1}^K w_k
 \left\{\frac12u^{(k)\top}A_{\tau,k}u^{(k)}
                         -g_\tau^{(k)\top}u^{(k)}\right\}\\
 &\hspace{2em}
 =\mu\left\{\frac12v^\top\widehat\Omega_{\tau,s}v
                                      -s_\tau^\top v\right\}+C_\tau,
 \end{aligned}
 \label{eq:hom-profile-source-aggregate}
\end{equation}
where $C_\tau$ does not depend on $v$ or $h$. To see this, the
constrained normal equations give
$A_{\tau,k}u^{(k)}-g_\tau^{(k)}=\ell$ for a common multiplier
$\ell$. Taking the weighted aggregate yields
$\ell=\widehat\Omega_{\tau,s}v-s_\tau$; substitution gives
\eqref{eq:hom-profile-source-aggregate}.
Combining this source loss with the target quadratic loss leaves,
up to a constant,
\[
 \frac12(v+h)^\top A_{\tau,0}(v+h)
       -g_\tau^{(0)\top}(v+h)
 +\mu\left\{\frac12v^\top\widehat\Omega_{\tau,s}v
                                      -s_\tau^\top v\right\}.
\]
For fixed $h$, the minimizing aggregate is therefore
\begin{equation}
 v_\tau(h)
 =\widehat H_\tau^{-1}
       \{g_\tau^{(0)}+\mu s_\tau-A_{\tau,0}h\}.
 \label{eq:hom-profile-aggregate-solution}
\end{equation}

Substituting \eqref{eq:hom-profile-aggregate-solution} identifies the
profile curvature and the two score components as
\begin{equation}
 \begin{aligned}
 M_\tau
 &=A_{\tau,0}
       -A_{\tau,0}\widehat H_\tau^{-1}A_{\tau,0},\\
 \zeta_{\tau,b}
 &=(I-A_{\tau,0}\widehat H_\tau^{-1})g_{\tau,b}^{(0)}
       -\mu A_{\tau,0}\widehat H_\tau^{-1}s_{\tau,b}\\
 &=\mathcal A_{\tau,0}g_{\tau,b}^{(0)}
       -\mathcal A_{\tau,s}s_{\tau,b},
 \qquad b\in\{e,r\}.
 \end{aligned}
 \label{eq:hom-profile-curvature-score}
\end{equation}
These are the coefficient matrices introduced in
Assumption~\ref{assump:joint-transfer}. For every
$h=\delta-\delta_\tau^*$, the resulting quadratic expansion is
\begin{equation}
 \mathcal L_{\mathrm{prof},\tau}(\delta)
       -\mathcal L_{\mathrm{prof},\tau}(\delta_\tau^*)
 =\frac12h^\top M_\tau h
       -\zeta_{\tau,e}^\top h-\zeta_{\tau,r}^\top h.
 \label{eq:hom-profile-expansion}
\end{equation}
Equivalently, its total linear score is
$\zeta_\tau=\zeta_{\tau,e}+\zeta_{\tau,r}$.
At the level of the quadratic part, the same profiling calculation gives
\begin{equation}
 h^\top M_\tau h
 =\inf_{v\in\mathbb R^q}
 \{(h+v)^\top A_{\tau,0}(h+v)
                  +\mu v^\top\widehat\Omega_{\tau,s}v\}.
 \label{eq:hom-profile-variational}
\end{equation}

Finally, let
\[
 \widehat h_\tau=\widehat\delta_\tau-\delta_\tau^*,\qquad
 \widehat v_\tau
 =\sum_{k=1}^K w_k
        (\widehat\beta_\tau^{(k)}-\beta_\tau^{(k)}).
\]
The fitted source aggregate satisfies
$\widehat v_\tau=v_\tau(\widehat h_\tau)$, and the target coefficient
perturbation equals $\widehat v_\tau+\widehat h_\tau$.
Thus \eqref{eq:hom-profile-aggregate-solution} gives the exact
reconstruction identity
\begin{equation}
 \begin{aligned}
 \widehat\beta_\tau^{(0)}-\beta_\tau^{(0)}
 & =\widehat H_\tau^{-1}
       \{g_\tau^{(0)}+\mu s_\tau\}
       +(I-\widehat H_\tau^{-1}A_{\tau,0})\widehat h_\tau\\
 & =\widehat H_\tau^{-1}
       \{g_\tau^{(0)}+\mu s_\tau\}
       +\mu\widehat H_\tau^{-1}
                   \widehat\Omega_{\tau,s}\widehat h_\tau.
 \end{aligned}
 \label{eq:hom-profile-target-reconstruction}
\end{equation}
The score bound and the one-step error bound below use
\eqref{eq:hom-profile-curvature-score}--\eqref{eq:hom-profile-expansion}
and \eqref{eq:hom-profile-target-reconstruction}, respectively.

\medskip
\noindent\textbf{Equivalent projection representation.}\par
For completeness, we verify the same profile expansion in observation space.
This calculation uses the $L_{\tau,k}$, $B$, $\alpha$ and
$\alpha_\tau^*$ already defined above.
We first derive the stacked least-squares representation. By definition,
\[
L_{\tau,0}(B\alpha+\delta)
=
\frac{1}{2n_0}
\left\|
y_\tau^{(0)}
-
Z_\tau^{(0)}B\alpha
-
Z_\tau^{(0)}\delta
\right\|_2^2,
\]
and, for each \(k=1,\ldots,K\),
\[
\mu w_kL_{\tau,k}(\beta^{(k)})
=
\frac{\mu w_k}{2n_k}
\left\|
y_\tau^{(k)}
-
Z_\tau^{(k)}\beta^{(k)}
\right\|_2^2.
\]

Define the stacked response vector
\[
\widetilde y_\tau
=
\begin{pmatrix}
n_0^{-1/2}y_\tau^{(0)}
\\
\sqrt{\mu w_1/n_1}\,y_\tau^{(1)}
\\
\vdots
\\
\sqrt{\mu w_K/n_K}\,y_\tau^{(K)}
\end{pmatrix}.
\]
Using the same block scaling, define the stacked regression-noise vector
\[
\widetilde e_\tau
=
\begin{pmatrix}
n_0^{-1/2}e_\tau^{(0)}
\\
\sqrt{\mu w_1/n_1}\,e_\tau^{(1)}
\\
\vdots
\\
\sqrt{\mu w_K/n_K}\,e_\tau^{(K)}
\end{pmatrix},
\]
and the stacked sieve-approximation vector
\[
\widetilde r_\tau
=
\begin{pmatrix}
n_0^{-1/2}r_\tau^{(0)}
\\
\sqrt{\mu w_1/n_1}\,r_\tau^{(1)}
\\
\vdots
\\
\sqrt{\mu w_K/n_K}\,r_\tau^{(K)}
\end{pmatrix}.
\]
Thus, the total stacked regression remainder is
\[
\widetilde\varepsilon_\tau
=
\widetilde e_\tau+\widetilde r_\tau.
\]

Define the correction design matrix
\[
\widetilde X_{\delta,\tau}
=
\begin{pmatrix}
n_0^{-1/2}Z_\tau^{(0)}
\\
0
\\
\vdots
\\
0
\end{pmatrix},
\]
and define the nuisance design matrix
\(\widetilde X_{\alpha,\tau}\) through
\[
\widetilde X_{\alpha,\tau}\alpha
=
\begin{pmatrix}
n_0^{-1/2}Z_\tau^{(0)}B\alpha
\\
\sqrt{\mu w_1/n_1}\,
Z_\tau^{(1)}\beta^{(1)}
\\
\vdots
\\
\sqrt{\mu w_K/n_K}\,
Z_\tau^{(K)}\beta^{(K)}
\end{pmatrix}.
\]
It follows directly that
\[
\ell_\tau(\delta,\alpha)
=
\frac12
\left\|
\widetilde y_\tau
-
\widetilde X_{\delta,\tau}\delta
-
\widetilde X_{\alpha,\tau}\alpha
\right\|_2^2.
\]

We next profile out the nuisance parameter \(\alpha\). For a fixed
\(\delta\), define
\[
v_\delta
=
\widetilde y_\tau
-
\widetilde X_{\delta,\tau}\delta.
\]
Then profiling over \(\alpha\) gives the ordinary least-squares problem
\[
\inf_\alpha
\frac12
\left\|
v_\delta
-
\widetilde X_{\alpha,\tau}\alpha
\right\|_2^2.
\]
Let \(P_{\alpha,\tau}\) denote the Euclidean projection matrix onto the column
space of \(\widetilde X_{\alpha,\tau}\). The minimum squared residual is
\[
\left\|
R_{\alpha,\tau}v_\delta
\right\|_2^2,
\qquad
R_{\alpha,\tau}
=
I-P_{\alpha,\tau}.
\]
Consequently,
\[
\mathcal L_{\mathrm{prof},\tau}(\delta)
=
\frac12
\left\|
R_{\alpha,\tau}
\left(
\widetilde y_\tau
-
\widetilde X_{\delta,\tau}\delta
\right)
\right\|_2^2.
\]

We now derive the quadratic expansion around the population correction.
Recall that the task-level regression representation is
\[
y_\tau^{(k)}
=
Z_\tau^{(k)}\beta_\tau^{(k)}
+
r_\tau^{(k)}
+
e_\tau^{(k)},
\qquad
k=0,1,\ldots,K.
\]
Moreover,
\[
B\alpha_\tau^*
=
\sum_{k=1}^K
w_k\beta_\tau^{(k)}
\]
and
\[
\beta_\tau^{(0)}
=
B\alpha_\tau^*
+
\delta_{\tau,\Delta}^{(0)}.
\]
Therefore, by the task-level regression representations and the stacked
definitions above,
\[
\widetilde y_\tau
=
\widetilde X_{\delta,\tau}
\delta_{\tau,\Delta}^{(0)}
+
\widetilde X_{\alpha,\tau}
\alpha_\tau^*
+
\widetilde e_\tau
+
\widetilde r_\tau.
\]

Since \(R_{\alpha,\tau}\) is the residual projection matrix associated with
\(\widetilde X_{\alpha,\tau}\),
\[
R_{\alpha,\tau}
\widetilde X_{\alpha,\tau}
\alpha_\tau^*
=
0.
\]
Let
\[
h
=
\delta-\delta_{\tau,\Delta}^{(0)}.
\]
It follows that
\[
\begin{aligned}
&
R_{\alpha,\tau}
\left(
\widetilde y_\tau
-
\widetilde X_{\delta,\tau}\delta
\right)
\\
={}&
R_{\alpha,\tau}
\left[
\widetilde e_\tau
+
\widetilde r_\tau
-
\widetilde X_{\delta,\tau}
\left(
\delta-\delta_{\tau,\Delta}^{(0)}
\right)
\right]
\\
={}&
R_{\alpha,\tau}\widetilde e_\tau
+
R_{\alpha,\tau}\widetilde r_\tau
-
R_{\alpha,\tau}\widetilde X_{\delta,\tau}h.
\end{aligned}
\]
Similarly,
\[
\begin{aligned}
&
R_{\alpha,\tau}
\left(
\widetilde y_\tau
-
\widetilde X_{\delta,\tau}
\delta_{\tau,\Delta}^{(0)}
\right)
\\
={}&
R_{\alpha,\tau}\widetilde e_\tau
+
R_{\alpha,\tau}\widetilde r_\tau.
\end{aligned}
\]

Hence,
\[
\begin{aligned}
&
\mathcal L_{\mathrm{prof},\tau}(\delta)
-
\mathcal L_{\mathrm{prof},\tau}
\left(
\delta_{\tau,\Delta}^{(0)}
\right)
\\
={}&
\frac12
\left\|
R_{\alpha,\tau}\widetilde e_\tau
+
R_{\alpha,\tau}\widetilde r_\tau
-
R_{\alpha,\tau}\widetilde X_{\delta,\tau}h
\right\|_2^2
\\
&-
\frac12
\left\|
R_{\alpha,\tau}\widetilde e_\tau
+
R_{\alpha,\tau}\widetilde r_\tau
\right\|_2^2
\\
={}&
\frac12
h^\top
\widetilde X_{\delta,\tau}^{\top}
R_{\alpha,\tau}^{\top}
R_{\alpha,\tau}
\widetilde X_{\delta,\tau}
h
\\
&-
\widetilde e_\tau^\top
R_{\alpha,\tau}^{\top}
R_{\alpha,\tau}
\widetilde X_{\delta,\tau}h
\\
&-
\widetilde r_\tau^\top
R_{\alpha,\tau}^{\top}
R_{\alpha,\tau}
\widetilde X_{\delta,\tau}h.
\end{aligned}
\]

Because \(R_{\alpha,\tau}\) is an orthogonal projection matrix,
\[
R_{\alpha,\tau}^{\top}
=
R_{\alpha,\tau},
\qquad
R_{\alpha,\tau}^2
=
R_{\alpha,\tau}.
\]
Therefore,
\[
R_{\alpha,\tau}^{\top}R_{\alpha,\tau}
=
R_{\alpha,\tau}.
\]
Substituting this identity into the preceding display yields
\[
\begin{aligned}
&
\mathcal L_{\mathrm{prof},\tau}(\delta)
-
\mathcal L_{\mathrm{prof},\tau}
\left(
\delta_{\tau,\Delta}^{(0)}
\right)
\\
={}&
\frac12
h^\top
\widetilde X_{\delta,\tau}^{\top}
R_{\alpha,\tau}
\widetilde X_{\delta,\tau}
h
\\
&-
\left(
\widetilde X_{\delta,\tau}^{\top}
R_{\alpha,\tau}
\widetilde e_\tau
\right)^\top h
\\
&-
\left(
\widetilde X_{\delta,\tau}^{\top}
R_{\alpha,\tau}
\widetilde r_\tau
\right)^\top h.
\end{aligned}
\]

By the definitions
\[
M_\tau
=
\widetilde X_{\delta,\tau}^{\top}
R_{\alpha,\tau}
\widetilde X_{\delta,\tau},
\]
\[
\zeta_{\tau,e}
=
\widetilde X_{\delta,\tau}^{\top}
R_{\alpha,\tau}
\widetilde e_\tau,
\qquad
\zeta_{\tau,r}
=
\widetilde X_{\delta,\tau}^{\top}
R_{\alpha,\tau}
\widetilde r_\tau,
\]
we obtain
\[
\begin{aligned}
&
\mathcal L_{\mathrm{prof},\tau}(\delta)
-
\mathcal L_{\mathrm{prof},\tau}
\left(
\delta_{\tau,\Delta}^{(0)}
\right)
\\
={}&
\frac12 h^\top M_\tau h
-
\zeta_{\tau,e}^\top h
-
\zeta_{\tau,r}^\top h.
\end{aligned}
\]
Equivalently, defining
\[
\zeta_\tau
=
\zeta_{\tau,e}
+
\zeta_{\tau,r},
\]
we have
\[
\mathcal L_{\mathrm{prof},\tau}(\delta)
-
\mathcal L_{\mathrm{prof},\tau}
\left(
\delta_{\tau,\Delta}^{(0)}
\right)
=
\frac12 h^\top M_\tau h
-
\zeta_\tau^\top h.
\]
Substituting back
\[
h
=
\delta-\delta_{\tau,\Delta}^{(0)}
\]
verifies the same quadratic expansion.

On $\mathcal E_{\Sigma,\tau}$, the projection form and the coefficient
form minimize the same quadratic loss. Their curvature and linear score
maps therefore coincide with \eqref{eq:hom-profile-curvature-score}.

\paragraph{Stochastic score bounds.}
We now control the stochastic scores. The infinity-norm bounds ensure
penalty domination, while the Euclidean bounds control the pooled score
contribution in target coefficient reconstruction.
\begin{lemma}[Concentration of the profiled stochastic score]
\label{lem:profiled-stochastic-score}
Fix an iteration $\tau$ and condition on the previous fitted iterates.
Use the raw and source-aggregate scores in
\eqref{eq:hom-profile-scores}.
Under the sample-splitting, reward-noise, and profile-stability conditions,
together with the conditions of Lemma~\ref{lem:uniform-empirical-gram},
for each $k=0,1,\ldots,K$,
\[
\left\|g_{\tau,e}^{(k)}\right\|_\infty
=
O_p\left(\sqrt{\frac{\log(q\vee2)}{n_k}}\right),
\]
and
\[
\left\|s_{\tau,e}\right\|_\infty
=
O_p\left(\sqrt{\frac{\log(q\vee2)}{n_s}}\right).
\]

The Euclidean scores also satisfy
\[
\left\|g_{\tau,e}^{(k)}\right\|_2
=O_p\!\left(\sqrt{\frac{q}{n_k}}\right),
\qquad
\left\|s_{\tau,e}\right\|_2
=O_p\!\left(\sqrt{\frac{q}{n_s}}\right).
\]
Moreover, the stochastic component of the profiled score satisfies
\[
\zeta_{\tau,e}
=
\mathcal A_{\tau,0}g_{\tau,e}^{(0)}
-
\mathcal A_{\tau,s}s_{\tau,e},
\]
and hence, under $\mu=n_s/n_0$,
\[
\left\|\zeta_{\tau,e}\right\|_\infty
=
O_p\left(
\sqrt{\frac{n_s\log(q\vee2)}{n_0n_{\mathrm{all}}}}
\right).
\]
If $q\to\infty$ and $c_\Delta$ is sufficiently large, then
\[
\left\|\zeta_{\tau,e}\right\|_\infty
\le
\frac{\lambda_\Delta}{2}
\]
with probability tending to one.
\end{lemma}

\begin{proof}[Proof of Lemma~\ref{lem:profiled-stochastic-score}]
We first control the task-specific raw scores. Fix a task $k$.
Let $\mathcal I_\tau^{(k)}$ be the indices of the complete
trajectories assigned to round $\tau$, ordered increasingly,
and let $\mathcal O_j^{(k)}$ denote the complete observed
trajectory indexed by $j$. Define exactly
\[
\mathcal G_{i,t}^{(k)}
:=
\mathcal F_{\tau-1}
\vee
\sigma\!\left(
\mathcal O_j^{(k)}:
j\in\mathcal I_\tau^{(k)},\ j<i
\right)
\vee
\mathcal H_{i,t}^{(k)}.
\]
The round-block independence and the independence of
trajectories within task $k$ extend the conditional reward
and transition conditions from $\mathcal H_{i,t}^{(k)}$ to
this sigma-field. In particular,
$\xi(X_{i,t}^{(k)},A_{i,t}^{(k)})$ is
$\mathcal G_{i,t}^{(k)}$-measurable.

Put
$V_{\tau-1}^{(k)}(x)
=\max_{a\in\mathcal A}\widehat Q_{\tau-1}^{(k)}(x,a)$
and
\[
Z_{i,t}^{(k),\tau}
=
V_{\tau-1}^{(k)}(X_{i,t+1}^{(k)})
-
(P^{(k)}V_{\tau-1}^{(k)})
(X_{i,t}^{(k)},A_{i,t}^{(k)}).
\]
Then
$e_{i,t}^{(k),\tau}
=\eta_{i,t}^{(k)}+\gamma Z_{i,t}^{(k),\tau}$
and
\[
\mathbb E[e_{i,t}^{(k),\tau}
\mid\mathcal G_{i,t}^{(k)}]=0.
\]
Let
\[
\mathcal E_{B,\tau-1}
=\left\{
\max_{0\le k\le K}\|\widehat Q_{\tau-1}^{(k)}\|_\infty\le B
\right\}\in\mathcal F_{\tau-1}.
\]
On $\mathcal E_{B,\tau-1}$, conditional Cauchy--Schwarz and
Hoeffding's lemma give
\[
\mathbb E\!\left[
\exp\{u e_{i,t}^{(k),\tau}\}
\mid\mathcal G_{i,t}^{(k)}
\right]
\le
\exp\{(\sigma_\eta^2+\gamma^2B^2)u^2\}
=
\exp\{\sigma_e^2u^2/2\},
\qquad
\sigma_e^2=2(\sigma_\eta^2+\gamma^2B^2).
\]

For coordinate $j\in[q]$,
\[
\left(g_{\tau,e}^{(k)}\right)_j
=
\frac{1}{n_k}
\sum_{(i,t)\in S_\tau^{(k)}}
\xi_j(X_{i,t}^{(k)},A_{i,t}^{(k)})
e_{i,t}^{(k),\tau}.
\]
Thus, on $\mathcal E_{B,\tau-1}$, the summands form a conditionally sub-Gaussian martingale-difference
array. On the empirical-Gram event,
\[
\frac{1}{n_k}
\sum_{(i,t)\in S_\tau^{(k)}}
\xi_j^2(X_{i,t}^{(k)},A_{i,t}^{(k)})
=
\left(\widehat\Sigma_\tau^{(k)}\right)_{jj}
\le C_{\mathrm G}.
\]
The exponential martingale inequality, together with a standard stopping
argument on this event, gives
\[
\mathbb P\left(
\left|\left(g_{\tau,e}^{(k)}\right)_j\right|>x,\,
\mathcal E_{\Sigma,\tau}\cap\mathcal E_{B,\tau-1}
\mid\mathcal F_{\tau-1}
\right)
\le
2\mathbf 1_{\mathcal E_{B,\tau-1}}\exp(-c n_kx^2)
\]
for some $c>0$. Taking a union bound over the $q$ coordinates
and using
$\mathbb P(\mathcal E_{\Sigma,\tau}^c)
+\mathbb P(\mathcal E_{B,\tau-1}^c)=o(1)$ yields
\[
\left\|g_{\tau,e}^{(k)}\right\|_\infty
=
O_p\left(\sqrt{\frac{\log(q\vee2)}{n_k}}\right).
\]

The same martingale differences give the Euclidean bound without a
coordinate union bound. On $\mathcal E_{B,\tau-1}$, the conditional
second moments of $e_{i,t}^{(k),\tau}$ are uniformly bounded by the mgf
bound above. Martingale orthogonality therefore gives
\[
\begin{aligned}
&\mathbf 1_{\mathcal E_{B,\tau-1}}
\mathbb E\!\left[
\left\|g_{\tau,e}^{(k)}\right\|_2^2
\mid\mathcal F_{\tau-1}\right]\\
&\qquad\lesssim
\frac{\mathbf 1_{\mathcal E_{B,\tau-1}}}{n_k^2}
\sum_{(i,t)\in S_\tau^{(k)}}
\mathbb E\!\left[
\left\|\xi(X_{i,t}^{(k)},A_{i,t}^{(k)})\right\|_2^2
\mid\mathcal F_{\tau-1}\right]
\lesssim
\mathbf 1_{\mathcal E_{B,\tau-1}}\frac{q}{n_k}.
\end{aligned}
\]
The last inequality uses the population Gram bound and the independence
of the current block from $\mathcal F_{\tau-1}$. Conditional Markov's
inequality and $\mathbb P(\mathcal E_{B,\tau-1}^c)=o(1)$ yield
\[
\left\|g_{\tau,e}^{(k)}\right\|_2
=O_p\!\left(\sqrt{\frac{q}{n_k}}\right).
\]

Next, by the definition of $\mathcal B_{\tau,k}$,
\[
s_{\tau,e}
=
\widehat\Omega_{\tau,s}
\sum_{k=1}^K
w_k
\left(\widehat\Sigma_\tau^{(k)}\right)^{-1}
g_{\tau,e}^{(k)}
=
\sum_{k=1}^K
w_k\mathcal B_{\tau,k}g_{\tau,e}^{(k)}.
\]
Assumption~\ref{assump:profile-stability} therefore implies
\[
\begin{aligned}
\left\|s_{\tau,e}\right\|_\infty
&\le
C_{\mathrm{prof}}
\sum_{k=1}^K
w_k\left\|g_{\tau,e}^{(k)}\right\|_\infty  \\
&=
O_p\left(
\frac{\sqrt{\log(q\vee2)}}{n_s}
\sum_{k=1}^K\sqrt{n_k}
\right)
=
O_p\left(\sqrt{\frac{\log(q\vee2)}{n_s}}\right),
\end{aligned}
\]
where the last equality uses
$\sum_{k=1}^K\sqrt{n_k}\le\sqrt{Kn_s}$ and the standing convention that
$K$ is fixed.

Since $K$ is fixed and
$\|\mathcal B_{\tau,k}\|_{\mathrm{op}}\lesssim1$ on
$\mathcal E_{\Sigma,\tau}$,
\[
\left\|
s_{\tau,e}
\right\|_2
=
O_p
\left(
\sqrt{\frac{q}{n_s}}
\right).
\]

For the effective stochastic score, use the identity in
\eqref{eq:hom-profile-curvature-score}.
Using Assumption~\ref{assump:profile-stability},
\[
\left\|\zeta_{\tau,e}\right\|_\infty
=
O_p\left[
\kappa_\mu
\left(
\sqrt{\frac{\log(q\vee2)}{n_0}}
+
\sqrt{\frac{\log(q\vee2)}{n_s}}
\right)
\right].
\]
Under $\mu=n_s/n_0$,
\[
\kappa_\mu=\frac{n_s}{n_{\mathrm{all}}},
\]
and
\[
\frac{\kappa_\mu}{\sqrt{n_0}}
\le
\sqrt{\frac{n_s}{n_0n_{\mathrm{all}}}},
\qquad
\frac{\kappa_\mu}{\sqrt{n_s}}
\le
\sqrt{\frac{n_s}{n_0n_{\mathrm{all}}}}.
\]
Therefore,
\[
\left\|\zeta_{\tau,e}\right\|_\infty
=
O_p\left(
\sqrt{\frac{n_s\log(q\vee2)}{n_0n_{\mathrm{all}}}}
\right).
\]
For $q\ge2$ and any fixed $A>0$, the coordinate tail bound above and a union
bound over tasks, coordinates, and the fixed number of iterations show that,
outside an event of probability at most
$2\Upsilon(K+1)q^{1-cA^2}+o(1)$,
\[
\left\|g_{\tau,e}^{(k)}\right\|_\infty
\le A\sqrt{\frac{\log q}{n_k}},
\qquad 0\le k\le K,\quad 1\le\tau\le\Upsilon.
\]
On the intersection of the event where the preceding bound
holds and the profile-stability event,
\[
\begin{aligned}
\max_{\tau\le\Upsilon}\left\|\zeta_{\tau,e}\right\|_\infty
&\le A C_{\mathrm{prof}}\kappa_\mu\sqrt{\log q}
\left(\frac{1}{\sqrt{n_0}}+
C_{\mathrm{prof}}\sum_{k=1}^K\frac{w_k}{\sqrt{n_k}}\right) \\
&\le A C_{\mathrm{prof}}(1+C_{\mathrm{prof}}\sqrt K)
\sqrt{\frac{n_s\log q}{n_0n_{\mathrm{all}}}}.
\end{aligned}
\]
Choose $A$ so that $cA^2>1$ and then choose
$c_\Delta\ge 2A C_{\mathrm{prof}}(1+C_{\mathrm{prof}}\sqrt K)$. Since
\[
\lambda_\Delta
=
c_\Delta
\sqrt{\frac{n_s\log q}{n_0n_{\mathrm{all}}}},
\]
it follows that
\[
\mathbb P\left(
\max_{\tau\le\Upsilon}\left\|\zeta_{\tau,e}\right\|_\infty
>\frac{\lambda_\Delta}{2}
\right)
\le 2\Upsilon(K+1)q^{1-cA^2}+o(1),
\]
which tends to zero when $q\to\infty$.
\end{proof}

\paragraph{One-step coefficient error.}
We combine the profile identities and score bounds with the sieve
remainder to obtain the one-step coefficient error below.
\begin{lemma}[One-step target coefficient error]
\label{lem:one-step-target-coefficient}
Suppose the sample-splitting, sieve-approximation, low-complexity correction,
and profile-stability conditions hold, together with the conditions of
Lemma~\ref{lem:uniform-empirical-gram}. Let
\[
\mu
=
\frac{n_s}{n_0},
\qquad
\lambda_\Delta
=
c_\Delta
\sqrt{
\frac{n_s\log q}
{n_0n_{\mathrm{all}}}
},
\]
where $c_\Delta>0$ is sufficiently large, and define
\[
\overline\lambda_\Delta
=
\frac{\lambda_\Delta}{\kappa_\mu},
\qquad
\kappa_\mu
=
\frac{\mu}{1+\mu}.
\]
Then, uniformly over $\tau=1,\ldots,\Upsilon$,
\[
\left\|
\widehat\delta_\tau
-
\delta_{\tau,\Delta}^{(0)}
\right\|_2
=
O_p
\left(
\Delta_{s,h_\Delta}
+
p^{-\kappa/d}
\right),
\]
where
\[
\Delta_{s,h_\Delta}
=
\sqrt{s}\,\overline\lambda_\Delta
\wedge
\sqrt{
h_\Delta\overline\lambda_\Delta
}
\wedge
h_\Delta.
\]
Moreover,
\[
\left\|
\widehat\beta_\tau^{(0)}
-
\beta_\tau^{(0)}
\right\|_2
=
O_p
\left(
\sqrt{
\frac{q}{n_{\mathrm{all}}}
}
+
\Delta_{s,h_\Delta}
+
p^{-\kappa/d}
\right).
\]
\end{lemma}

\begin{proof}[Proof of Lemma~\ref{lem:one-step-target-coefficient}]
Fix an iteration $\tau$ and condition on the previous fitted iterates.
Use the notation of the direct profiling calculation, and write
$a_p=p^{-\kappa/d}$.

\noindent\textbf{Step 1: Profile curvature and sieve approximation.}\par
By Lemma~\ref{lem:uniform-empirical-gram},
$\mathbb P(\mathcal E_{\Sigma,\tau})\longrightarrow1$.
On this event, the empirical Gram matrices and
$\widehat\Omega_{\tau,s}$ have eigenvalues bounded above and below.
The variational identity \eqref{eq:hom-profile-variational} consequently gives
\[
c_M\kappa_\mu\|h\|_2^2
\le h^\top M_\tau h
\le C_M\kappa_\mu\|h\|_2^2,
\qquad \kappa_\mu=\frac{\mu}{1+\mu},
\]
for constants $0<c_M<C_M<\infty$.

For the approximation scores in \eqref{eq:hom-profile-scores},
for every $v\in\mathbb R^q$ with $\|v\|_2=1$,
\[
\begin{aligned}
v^\top
A_{\tau,k}^{-1/2}g_{\tau,r}^{(k)}
&=
\frac{1}{n_k}
\left(
Z_\tau^{(k)}A_{\tau,k}^{-1/2}v
\right)^\top
r_\tau^{(k)}
\\
&\le
\left[
\frac{1}{n_k}
\left\|
Z_\tau^{(k)}A_{\tau,k}^{-1/2}v
\right\|_2^2
\right]^{1/2}
\frac{\|r_\tau^{(k)}\|_2}{\sqrt{n_k}}
\\
&\lesssim
a_p.
\end{aligned}
\]
Consequently,
\[
\left\|
g_{\tau,r}^{(k)}
\right\|_2
\lesssim
a_p.
\]

The empirical-Gram bounds imply
\[
\|s_{\tau,r}\|_2
\lesssim
a_p.
\]

For the approximation-score identity in
\eqref{eq:hom-profile-curvature-score}, note that
\[
I-A_{\tau,0}\widehat H_\tau^{-1}
=
\mu\widehat\Omega_{\tau,s}\widehat H_\tau^{-1},
\]
and
\[
\left\|
\widehat H_\tau^{-1}
\right\|_{\mathrm{op}}
\lesssim
\frac{1}{1+\mu}.
\]
It follows that
\[
\left\|
\zeta_{\tau,r}
\right\|_2
\lesssim
\kappa_\mu a_p,
\]
and hence
\[
\left|
\zeta_{\tau,r}^\top h
\right|
\lesssim
\kappa_\mu a_p\|h\|_2.
\]

\noindent\textbf{Step 2: Basic inequality.}\par

By \eqref{eq:hom-profile-minimizer} and
\eqref{eq:hom-profile-expansion}, the optimality of
$\widehat\delta_\tau$ gives
\[
\begin{aligned}
\frac12
\widehat h_\tau^\top M_\tau\widehat h_\tau
\le{}&
\zeta_{\tau,e}^\top\widehat h_\tau
+
\zeta_{\tau,r}^\top\widehat h_\tau
\\
&+
\lambda_\Delta
\left(
\|\delta_\tau^*\|_1
-
\|\delta_\tau^*+\widehat h_\tau\|_1
\right).
\end{aligned}
\]

We first treat $q\to\infty$, using the stochastic-score event from
Lemma~\ref{lem:profiled-stochastic-score},
\[
\left\|
\zeta_{\tau,e}
\right\|_\infty
\le
\frac{\lambda_\Delta}{2}.
\]
Combining this event with the bounds from Step 1 yields
\begin{equation}
\begin{aligned}
c\kappa_\mu
\|\widehat h_\tau\|_2^2
\le{}&
\frac{\lambda_\Delta}{2}
\|\widehat h_\tau\|_1
+
\lambda_\Delta
\left(
\|\delta_\tau^*\|_1
-
\|\delta_\tau^*+\widehat h_\tau\|_1
\right)
\\
&+
C\kappa_\mu a_p
\|\widehat h_\tau\|_2.
\end{aligned}
\label{eq:app-a1}
\end{equation}

If $\|\widehat h_\tau\|_2\le C_0a_p$ for a sufficiently large constant
$C_0$, the desired approximation bound already holds. Otherwise, the final
term in \eqref{eq:app-a1} can be absorbed into its left-hand side. Thus it remains to
analyze
\begin{equation}
c\kappa_\mu
\|\widehat h_\tau\|_2^2
\le
\frac{\lambda_\Delta}{2}
\|\widehat h_\tau\|_1
+
\lambda_\Delta
\left(
\|\delta_\tau^*\|_1
-
\|\delta_\tau^*+\widehat h_\tau\|_1
\right).
\label{eq:app-a2}
\end{equation}

\noindent\textbf{Step 3: Adaptive discrepancy bounds.}\par

Let
\[
S_\tau
=
\operatorname{supp}(\delta_\tau^*).
\]
Since $\delta_{\tau,S_\tau^c}^*=0$,
\[
\|\delta_\tau^*\|_1
-
\|\delta_\tau^*+\widehat h_\tau\|_1
\le
\|\widehat h_{\tau,S_\tau}\|_1
-
\|\widehat h_{\tau,S_\tau^c}\|_1.
\]
Substituting this inequality into \eqref{eq:app-a2} and dropping the nonpositive term gives
\[
c\kappa_\mu
\|\widehat h_\tau\|_2^2
\le
\frac{3\lambda_\Delta}{2}
\|\widehat h_{\tau,S_\tau}\|_1
\le
\frac{3\lambda_\Delta}{2}
\sqrt{s}\,
\|\widehat h_\tau\|_2.
\]
Therefore,
\begin{equation}
\|\widehat h_\tau\|_2
\lesssim
\sqrt{s}\,
\frac{\lambda_\Delta}{\kappa_\mu}.
\label{eq:app-a3}
\end{equation}

Alternatively, the reverse triangle inequality and
$\|\delta_\tau^*\|_1\le h_\Delta$ imply
\[
\|\delta_\tau^*\|_1
-
\|\delta_\tau^*+\widehat h_\tau\|_1
\le
2h_\Delta-\|\widehat h_\tau\|_1.
\]
Using this inequality in \eqref{eq:app-a2} and dropping the resulting nonpositive
$\ell_1$ term gives
\[
c\kappa_\mu
\|\widehat h_\tau\|_2^2
\le
2\lambda_\Delta h_\Delta,
\]
and hence
\begin{equation}
\|\widehat h_\tau\|_2
\lesssim
\sqrt{
h_\Delta
\frac{\lambda_\Delta}{\kappa_\mu}
}.
\end{equation}
If instead we retain the negative term, then
\[
c\kappa_\mu
\|\widehat h_\tau\|_2^2
+
\frac{\lambda_\Delta}{2}
\|\widehat h_\tau\|_1
\le
2\lambda_\Delta h_\Delta.
\]
Dropping the quadratic term yields
\begin{equation}
\|\widehat h_\tau\|_2
\le
\|\widehat h_\tau\|_1
\lesssim
h_\Delta.
\label{eq:app-a5}
\end{equation}

Combining \eqref{eq:app-a3}--\eqref{eq:app-a5} with the case
$\|\widehat h_\tau\|_2\lesssim a_p$ gives
\[
\left\|
\widehat\delta_\tau-\delta_\tau^*
\right\|_2
=
O_p
\left(
\sqrt{s}\,\overline\lambda_\Delta
\wedge
\sqrt{
h_\Delta\overline\lambda_\Delta
}
\wedge
h_\Delta
+
a_p
\right),
\]
where
\[
\overline\lambda_\Delta
=
\frac{\lambda_\Delta}{\kappa_\mu}.
\]
Equivalently,
\begin{equation}
\left\|
\widehat\delta_\tau-\delta_\tau^*
\right\|_2
=
O_p
\left(
\Delta_{s,h_\Delta}
+
p^{-\kappa/d}
\right).
\label{eq:app-a6}
\end{equation}

\noindent\textbf{Step 4: Reconstruction of the target coefficient.}\par
The exact identity \eqref{eq:hom-profile-target-reconstruction}
separates the pooled score contribution from the correction contribution.
By the empirical-Gram bounds,
\[
\left\|
\widehat H_\tau^{-1}
\right\|_{\mathrm{op}}
\lesssim
\frac{1}{1+\mu}.
\]
Together with the Euclidean score bounds in
Lemma~\ref{lem:profiled-stochastic-score}, this gives
\begin{equation}
\begin{aligned}
\left\|
\widehat H_\tau^{-1}
\left[
g_{\tau,e}^{(0)}
+
\mu s_{\tau,e}
\right]
\right\|_2
&=
O_p\left[
\frac{1}{1+\mu}
\left(
\sqrt{\frac{q}{n_0}}
+
\mu\sqrt{\frac{q}{n_s}}
\right)
\right]
\\
&=
O_p\left(
\sqrt{\frac{q}{n_{\mathrm{all}}}}
\right),
\end{aligned}
\end{equation}
where the last equality uses $\mu=n_s/n_0$.

The approximation-score bounds from Step 1 similarly imply
\begin{equation}
\left\|
\widehat H_\tau^{-1}
\left[
g_{\tau,r}^{(0)}
+
\mu s_{\tau,r}
\right]
\right\|_2
\lesssim
a_p.
\end{equation}
Finally,
\begin{equation}
\left\|
\mu\widehat H_\tau^{-1}
\widehat\Omega_{\tau,s}
\right\|_{\mathrm{op}}
\lesssim1.
\label{eq:app-a10}
\end{equation}
Combining \eqref{eq:app-a6},
\eqref{eq:hom-profile-target-reconstruction}, and the preceding
stochastic, approximation, and correction-map bounds yields
\[
\left\|
\widehat\beta_\tau^{(0)}
-
\beta_\tau^{(0)}
\right\|_2
=
O_p
\left(
\sqrt{\frac{q}{n_{\mathrm{all}}}}
+
\Delta_{s,h_\Delta}
+
p^{-\kappa/d}
\right).
\]

\noindent\textbf{Probability bookkeeping.}\par
If $q$ is bounded, the optimality inequality in Step 2, Step 1, and
$\|\delta_\tau^*\|_1-\|\delta_\tau^*+\widehat h_\tau\|_1
\le\sqrt q\|\widehat h_\tau\|_2$ give
\[
\|\widehat h_\tau\|_2
\lesssim a_p+
\frac{\sqrt q}{\kappa_\mu}
\left(\|\zeta_{\tau,e}\|_\infty+\lambda_\Delta\right).
\]
By Lemma~\ref{lem:profiled-stochastic-score} and the definition of
$\lambda_\Delta$, the second term is $o_p(1)$ for bounded $q$, since
$n_{\min}\to\infty$. As $q=mp$ is bounded, $a_p=p^{-\kappa/d}$ is bounded
below, so $\|\widehat h_\tau\|_2=O_p(a_p)$ in this case.

Thus \eqref{eq:app-a6} also holds for bounded $q$, and the same
reconstruction bounds in Step 4 give the target coefficient bound.
The preceding argument proves \eqref{eq:app-a6} when $q\to\infty$; the bounded-$q$
argument above proves it when $q$ remains bounded.
The same bound holds for sequences on which $q$ does not converge: split at a
fixed $q_0$. The bounded-$q$ argument applies for $q<q_0$, while for
$q\ge q_0$ the score-event failure probability is at most
$2\Upsilon(K+1)q_0^{1-cA^2}+o(1)$, which can be made arbitrarily small by
increasing $q_0$.

The reconstruction bounds apply to both parts of this split.
Since the number of FQI iterations is fixed, the conclusion holds uniformly
over $\tau=1,\ldots,\Upsilon$ by a finite union bound. This completes the
proof.
\end{proof}

\paragraph{Proof of Theorem~\ref{thm:l1-joint-transfer-fqi}.}
\begin{proof}
\noindent\textbf{Step 1: From coefficient error to the Bellman residual.}\par
Write
\[
 \mathfrak r_n
 =p^{-\kappa/d}+\sqrt{q/n_{\mathrm{all}}}+\Delta_{s,h_\Delta}.
\]
Lemma~\ref{lem:one-step-target-coefficient} gives
\[
 \left\|\widehat\beta_\tau^{(0)}-\beta_\tau^{(0)}\right\|_2
 =O_p(\mathfrak r_n),
\]
uniformly over $\tau=1,\ldots,\Upsilon$. We translate this bound into
control of the target Bellman residual.

Let $\mathfrak D_0$ denote the collection of state-action distributions
generated from the target initial distribution $\nu^{(0)}$ by arbitrary
finite sequences of target-task Markov policies, allowing arbitrary
current-action choices at the final state, together with their countable
convex mixtures, including discounted occupancy distributions. For a
measurable function $g$, write
\[
 \|g\|_{L_1(\nu)}=\mathbb E_\nu[|g(X,A)|],
 \qquad \nu\in\mathfrak D_0.
\]
These integrals are evaluated with the fitted functions held fixed.

By Assumption~\ref{assump:ergodicity}, the target initial-state density is
uniformly bounded. Moreover, under the standard definition of the
H\"older ball, Assumption~\ref{assump:smoothness} implies that the
transition density is uniformly bounded. Hence, the state marginal of
every distribution in $\mathfrak D_0$ has a uniformly bounded density.
In addition, Lemma~E.1 of the supplementary material to
\citet{Shi2022ValueInference} gives the following standard property for
B-spline and wavelet bases:
\[
 \lambda_{\max}\left(\int_{\mathcal X}
                 \phi(x)\phi^\top(x)\,dx\right)\lesssim1.
\]
It follows that, uniformly over $\nu\in\mathfrak D_0$,
\begin{equation}
 \mathbb E_\nu\big[\{\xi^\top(X,A)u\}^2\big]
 \lesssim\|u\|_2^2,\qquad u\in\mathbb R^q.
 \label{eq:app-a11}
\end{equation}
Indeed, writing $u=(u_1^\top,\ldots,u_m^\top)^\top$ with
$u_a\in\mathbb R^p$, the left-hand side of \eqref{eq:app-a11} is bounded
by a constant multiple of
\[
 \sum_{a=1}^m u_a^\top
       \left(\int_{\mathcal X}\phi(x)\phi^\top(x)\,dx\right)u_a
 \lesssim\sum_{a=1}^m\|u_a\|_2^2=\|u\|_2^2.
\]

For iteration $\tau$, define the target Bellman residual
\[
 d_\tau=\mathcal T^{(0)}\widehat Q_{\tau-1}^{(0)}
                              -\widehat Q_\tau^{(0)}.
\]
Since $f_\tau^{(0)}=\mathcal T^{(0)}\widehat Q_{\tau-1}^{(0)}$ and
$\widehat Q_\tau^{(0)}(x,a)=\xi^\top(x,a)\widehat\beta_\tau^{(0)}$,
\[
 \begin{split}
 d_\tau(x,a)={}&f_\tau^{(0)}(x,a)-\xi^\top(x,a)\beta_\tau^{(0)}\\
 &+\xi^\top(x,a)\{\beta_\tau^{(0)}-\widehat\beta_\tau^{(0)}\}.
 \end{split}
\]
Consequently, the sieve approximation bound, \eqref{eq:app-a11}, and
Cauchy--Schwarz imply
\begin{equation}
 \begin{split}
 \sup_{\nu\in\mathfrak D_0}\|d_\tau\|_{L_1(\nu)}
 &\lesssim p^{-\kappa/d}
                  +\|\widehat\beta_\tau^{(0)}-\beta_\tau^{(0)}\|_2\\
 &=O_p(\mathfrak r_n),
 \end{split}
 \label{eq:app-a12}
\end{equation}
uniformly over $\tau=1,\ldots,\Upsilon$.

\medskip
\noindent\textbf{Step 2: FQI error propagation and initialization.}\par
Write $Q^*=Q^{*,(0)}$. For any policy $\omega$ and measurable function
$g:\mathcal X\times\mathcal A\to\mathbb R$, define
\[
 (P^\omega g)(x,a)
 =\mathbb E\big[g(X',A')\mid
        X'\sim P(\cdot\mid x,a),\ A'\sim\omega(\cdot\mid X')\big].
\]
The reward bound and the assumed initialization give
\[
 \|Q^*-\widehat Q_0^{(0)}\|_\infty
 \le\|Q^*\|_\infty+\|\widehat Q_0^{(0)}\|_\infty
 \le\frac{2R_{\max}}{1-\gamma}.
\]
Each $P^\omega$ preserves uniform bounds. Following the error-propagation
argument in Lemma~C.2 of \citet{Fan2020DeepQLearning}, and taking absolute
values and suprema over Markov policy sequences, we therefore obtain
\begin{equation}
 \begin{split}
 |Q^*(x,a)-\widehat Q_\Upsilon^{(0)}(x,a)|
 \le{}&\sum_{\tau=1}^\Upsilon\gamma^{\Upsilon-\tau}
   \sup_{\omega_1,\ldots,\omega_{\Upsilon-\tau}}
   \big(P^{\omega_{\Upsilon-\tau}}\cdots P^{\omega_1}|d_\tau|\big)(x,a)\\
 &+\frac{2\gamma^\Upsilon R_{\max}}{1-\gamma}.
 \end{split}
 \label{eq:app-homogeneous-fqi-propagation}
\end{equation}
Here a product of zero transition operators is the identity, and the
supremum over an empty policy sequence leaves $|d_\Upsilon|$ unchanged.

For a nonnegative terminal function, backward maximization over the
finite action set, with a fixed tie-breaking rule, constructs measurable
Markov policies attaining each pointwise supremum in
\eqref{eq:app-homogeneous-fqi-propagation}. Appending such a policy
sequence preserves $\mathfrak D_0$. Thus, integrating each propagated
residual against any $\nu\in\mathfrak D_0$ is bounded by
$\sup_{\nu'\in\mathfrak D_0}\|d_\tau\|_{L_1(\nu')}$.
Using \eqref{eq:app-a12} and summing the geometric series gives
\begin{equation}
 \begin{split}
 \sup_{\nu\in\mathfrak D_0}
       \|Q^*-\widehat Q_\Upsilon^{(0)}\|_{L_1(\nu)}
 &\le O_p\left(\mathfrak r_n
                      \sum_{j=0}^{\Upsilon-1}\gamma^j\right)
                 +\frac{2\gamma^\Upsilon R_{\max}}{1-\gamma}\\
 &=O_p\left(\frac{\mathfrak r_n}{1-\gamma}
                  +\frac{\gamma^\Upsilon R_{\max}}{1-\gamma}\right).
 \end{split}
 \label{eq:app-a13}
\end{equation}

\medskip
\noindent\textbf{Step 3: From the greedy policy to value loss.}\par
Let
\[
 \widehat\pi_\Upsilon^{(0)}(x)
 \in\arg\max_{a\in\mathcal A}\widehat Q_\Upsilon^{(0)}(x,a).
\]
The greedy property implies, for every $x$,
\[
 \begin{split}
 &Q^*\big(x,\pi^{*,(0)}(x)\big)
       -Q^*\big(x,\widehat\pi_\Upsilon^{(0)}(x)\big)\\
 &\quad\le
   \big|Q^*-\widehat Q_\Upsilon^{(0)}\big|
                      \big(x,\pi^{*,(0)}(x)\big)
   +\big|Q^*-\widehat Q_\Upsilon^{(0)}\big|
                      \big(x,\widehat\pi_\Upsilon^{(0)}(x)\big).
 \end{split}
\]
The normalized discounted state occupancy of the learned policy is a
countable mixture of reachable state distributions. Its current-action
lifts under $\pi^{*,(0)}$ and $\widehat\pi_\Upsilon^{(0)}$ both belong
to $\mathfrak D_0$. Applying the standard performance-difference
argument under this occupancy, and then \eqref{eq:app-a13}, yields
\[
 \begin{split}
 v^{\pi^{*,(0)},(0)}-v^{\widehat\pi_\Upsilon^{(0)},(0)}
 &\le\frac{2}{1-\gamma}\sup_{\nu\in\mathfrak D_0}
              \|Q^*-\widehat Q_\Upsilon^{(0)}\|_{L_1(\nu)}\\
 &=O_p\left(\frac{\mathfrak r_n}{(1-\gamma)^2}
             +\frac{\gamma^\Upsilon R_{\max}}{(1-\gamma)^2}\right)\\
 &=\frac{1}{(1-\gamma)^2}O_p\left(
      p^{-\kappa/d}+\sqrt{q/n_{\mathrm{all}}}
        +\Delta_{s,h_\Delta}+\gamma^\Upsilon R_{\max}\right).
 \end{split}
\]
This establishes the stated value-loss bound and completes the proof of
Theorem~\ref{thm:l1-joint-transfer-fqi}.
\end{proof}

\paragraph{Proof of Theorem~\ref{thm:heterogeneous-l1-joint-transfer-fqi}}

We now consider the transition-heterogeneous setting. The reward
functions, behavior policies, initial-state distributions, and
population sieve design matrices may vary across tasks, as in the
transition-homogeneous setting. Here, the transition kernels
$P^{(k)}$ may vary as well.

For each task \(k\in\{0\}\cup[K]\), define its Bellman optimality operator by
\[
\left(
\mathcal T^{(k)}Q
\right)(x,a)
=
r^{(k)}(x,a)
+
\gamma
\int_{\mathcal X}
\max_{a'\in\mathcal A}
Q(x',a')
\rho^{(k)}(x'\mid x,a)\,dx'.
\]

For a target-task policy \(\omega\), define
\[
\left(
P^{(0),\omega}g
\right)(x,a)
=
\mathbb E
\left[
g(X',A')
\mid
X'\sim P^{(0)}(\cdot\mid x,a),
\;
A'\sim\omega(\cdot\mid X')
\right].
\]

Allowing source transition kernels to differ from the target
kernel does not change the FQI error-propagation argument for
the target estimator. This argument involves only the target
Bellman operator \(\mathcal T^{(0)}\) and the target transition
kernel \(P^{(0)}\). Therefore, with the convention that a product
of zero transition operators is the identity operator,
\begin{equation}
\begin{aligned}
\left|
Q^{*,(0)}
-
\widehat Q_\Upsilon^{(0)}
\right|
\le{}&
\sum_{\tau=0}^{\Upsilon-1}
\gamma^{\Upsilon-1-\tau}
\sup_{\omega_1,\ldots,\omega_{\Upsilon-1-\tau}}
\left(
P^{(0),\omega_{\Upsilon-1-\tau}}
\cdots
P^{(0),\omega_1}
\right)
\\
&\qquad\qquad\qquad\qquad
\left|
\mathcal T^{(0)}
\widehat Q_\tau^{(0)}
-
\widehat Q_{\tau+1}^{(0)}
\right|
+
\frac{
2\gamma^\Upsilon R_{\max}
}{
1-\gamma
}.
\end{aligned}
\label{eq:app-h1}
\end{equation}
Consequently, it suffices to control the one-step target Bellman residual
\[
\left|
\mathcal T^{(0)}
\widehat Q_{\tau-1}^{(0)}
-
\widehat Q_\tau^{(0)}
\right|
\]
uniformly over \(\tau=1,\ldots,\Upsilon\).

Fix an iteration $\tau$, and let $\mathcal F_{\tau-1}$ be the
sigma-field generated by the initial fitted functions and the
data blocks $\mathcal D_1,\ldots,\mathcal D_{\tau-1}$.
The round block $\mathcal D_\tau$ is independent of
$\mathcal F_{\tau-1}$.

For every task \(k\in\{0\}\cup[K]\), the pseudo-response is
\[
Y_{i,t}^{(k),\tau}
=
R_{i,t}^{(k)}
+
\gamma
\max_{a'\in\mathcal A}
\widehat Q_{\tau-1}^{(k)}
\left(
X_{i,t+1}^{(k)},a'
\right),
\qquad
(i,t)\in S_\tau^{(k)}.
\]
Its conditional expectation is
\[
\begin{aligned}
&\mathbb E
\left[
Y_{i,t}^{(k),\tau}
\mid
X_{i,t}^{(k)},A_{i,t}^{(k)},
\mathcal F_{\tau-1}
\right]
\\
&\qquad=
\left(
\mathcal T^{(k)}
\widehat Q_{\tau-1}^{(k)}
\right)
\left(
X_{i,t}^{(k)},A_{i,t}^{(k)}
\right).
\end{aligned}
\]
Define the task-specific population Bellman regression function
\[
f_\tau^{(k)}
=
\mathcal T^{(k)}
\widehat Q_{\tau-1}^{(k)}.
\]
Then
\[
Y_{i,t}^{(k),\tau}
=
f_\tau^{(k)}
\left(
X_{i,t}^{(k)},A_{i,t}^{(k)}
\right)
+
e_{i,t}^{(k),\tau},
\]
where
\[
\mathbb E
\left[
e_{i,t}^{(k),\tau}
\mid
X_{i,t}^{(k)},A_{i,t}^{(k)},
\mathcal F_{\tau-1}
\right]
=
0.
\]
Under the reward-noise condition and the uniform boundedness of the previous
fitted iterates, \(e_{i,t}^{(k),\tau}\) is conditionally sub-Gaussian.

By the approximation property in Assumption~\ref{assump:smoothness},
the coefficients defined above for each task \(k\), namely
\[
\beta_r^{(k)}
\in
\mathbb R^q
\]
and
\[
\left\{
\beta_\rho^{(k)}(x')
\in
\mathbb R^q:
x'\in\mathcal X
\right\}
\]
satisfy
\[
\sup_{x,a,k}
\left|
r^{(k)}(x,a)
-
\xi^\top(x,a)\beta_r^{(k)}
\right|
\lesssim
p^{-\kappa/d},
\]
and
\[
\sup_{x,a,x',k}
\left|
\rho^{(k)}(x'\mid x,a)
-
\xi^\top(x,a)
\beta_\rho^{(k)}(x')
\right|
\lesssim
p^{-\kappa/d}.
\]

Let
\[
V_{\tau-1}^{(k)}(x')
=
\max_{a'\in\mathcal A}
\widehat Q_{\tau-1}^{(k)}(x',a'),
\]
and define the task-specific one-step Bellman coefficient
\begin{equation}
\beta_\tau^{(k)}
=
\beta_r^{(k)}
+
\gamma
\int_{\mathcal X}
V_{\tau-1}^{(k)}(x')
\beta_\rho^{(k)}(x')\,dx'.
\label{eq:app-h2}
\end{equation}
By the definition of \(f_\tau^{(k)}\),
\[
f_\tau^{(k)}(x,a)
=
r^{(k)}(x,a)
+
\gamma
\int_{\mathcal X}
V_{\tau-1}^{(k)}(x')
\rho^{(k)}(x'\mid x,a)\,dx'.
\]
The uniform boundedness of \(V_{\tau-1}^{(k)}\), together with the sieve
approximation bounds above, implies
\begin{equation}
\sup_{x,a,k}
\left|
f_\tau^{(k)}(x,a)
-
\xi^\top(x,a)\beta_\tau^{(k)}
\right|
\lesssim
p^{-\kappa/d}.
\end{equation}

Let \(Z_\tau^{(k)}\in\mathbb R^{n_k\times q}\) be the design matrix whose
rows are
\[
\xi^\top
\left(
X_{i,t}^{(k)},A_{i,t}^{(k)}
\right),
\qquad
(i,t)\in S_\tau^{(k)}.
\]
Let \(y_\tau^{(k)}\) collect the pseudo-responses
\(Y_{i,t}^{(k),\tau}\), and let \(e_\tau^{(k)}\) collect the corresponding
regression noise terms. Define the sample sieve-approximation remainder by
\[
r_\tau^{(k)}
=
f_\tau^{(k)}
-
Z_\tau^{(k)}
\beta_\tau^{(k)},
\]
where \(f_\tau^{(k)}\) on the right-hand side denotes the vector obtained by
evaluating the regression function at the sample design points.

The task-specific regression representation is therefore
\begin{equation}
y_\tau^{(k)}
=
Z_\tau^{(k)}
\beta_\tau^{(k)}
+
r_\tau^{(k)}
+
e_\tau^{(k)},
\qquad
k=0,1,\ldots,K,
\end{equation}
with
\begin{equation}
\left\|
r_\tau^{(k)}
\right\|_\infty
\lesssim
p^{-\kappa/d}.
\end{equation}

In this setting, the population Bellman coefficients
\(\beta_\tau^{(k)}\), population design matrices
\(\Sigma^{(k)}\), and associated discrepancy parameters may
differ across tasks. The algebraic form of the joint regression
problem remains unchanged. The remaining task
is to characterize the target-source aggregate correction
\[
\delta_{\tau,\Delta}^{(0)}
=
\beta_\tau^{(0)}
-
\sum_{k=1}^K
w_k\beta_\tau^{(k)}
\]
under heterogeneous rewards and transition kernels, after which the profiled
joint-update analysis can be applied.

To quantify the primitive heterogeneity between the target task and the
source tasks, define, for each \(k\in[K]\),
\[
h_k
=
\left\|
\beta_r^{(k)}-\beta_r^{(0)}
\right\|_1
+
\int_{\mathcal X}
\left\|
\beta_\rho^{(k)}(x')
-
\beta_\rho^{(0)}(x')
\right\|_1
\,dx',
\]
and define the source-weighted primitive discrepancy
\[
h_{\mathrm w}
=
\sum_{k=1}^K
w_k h_k.
\]

We also define the weighted discrepancy between the fitted task-specific
coefficients at iteration \(\tau\) by
\[
d_\tau
=
\sum_{k=1}^K
w_k
\left\|
\widehat\beta_\tau^{(k)}
-
\widehat\beta_\tau^{(0)}
\right\|_2.
\]

The following lemma relates the target-source aggregate Bellman correction to
the primitive task heterogeneity and the discrepancy inherited from the
previous FQI iteration.

\begin{lemma}[Weighted Bellman discrepancy under transition heterogeneity]
\label{lem:weighted-bellman-discrepancy}
Suppose the conditions of
Theorem~\ref{thm:heterogeneous-l1-joint-transfer-fqi} hold, including the
bounded-iterate condition and the transition-coefficient stability condition
\eqref{eq:transition-coefficient-stability}. Then, uniformly over \(\tau=1,\ldots,\Upsilon\),
\begin{equation}
\left\|
\delta_{\tau,\Delta}^{(0)}
\right\|_1
\lesssim
h_{\mathrm w}
+
\gamma d_{\tau-1}.
\label{eq:app-h7}
\end{equation}
\end{lemma}

Inequality \eqref{eq:app-h7} shows that the aggregate Bellman correction is controlled by
the primitive weighted heterogeneity \(h_{\mathrm w}\), together with the
task discrepancy inherited from the preceding FQI iteration. To close this
recursion, define
\[
E_{\tau,0}
=
\left\|
\widehat\beta_\tau^{(0)}
-
\beta_\tau^{(0)}
\right\|_2
\]
and
\[
E_{\tau,s}
=
\sum_{k=1}^K
w_k
\left\|
\widehat\beta_\tau^{(k)}
-
\beta_\tau^{(k)}
\right\|_2.
\]
By adding and subtracting the corresponding population Bellman coefficients,
we have
\begin{equation}
\begin{aligned}
d_\tau
&=
\sum_{k=1}^K
w_k
\left\|
\widehat\beta_\tau^{(k)}
-
\widehat\beta_\tau^{(0)}
\right\|_2
\\
&\le
E_{\tau,s}
+
E_{\tau,0}
+
\sum_{k=1}^K
w_k
\left\|
\beta_\tau^{(k)}
-
\beta_\tau^{(0)}
\right\|_2
\\
&\lesssim
E_{\tau,s}
+
E_{\tau,0}
+
h_{\mathrm w}
+
\gamma d_{\tau-1},
\end{aligned}
\label{eq:app-h8}
\end{equation}
where the last inequality follows from the argument underlying
Lemma~\ref{lem:weighted-bellman-discrepancy} and the inequality
\(\|u\|_2\le\|u\|_1\).

The target estimation error \(E_{\tau,0}\) is controlled by
Lemma~\ref{lem:one-step-target-coefficient}. It remains to derive a
corresponding reconstruction bound for the weighted source estimation error
\(E_{\tau,s}\). The next lemma provides this bound. The proofs of the
auxiliary lemmas are collected after their statements.

\begin{lemma}[Weighted source-coefficient reconstruction]
\label{lem:weighted-source-reconstruction}
Fix an iteration \(\tau\), and condition on the previous fitted iterates.
Define
\[
g_\tau^{(k)}
=
\frac{1}{n_k}
\left(
Z_\tau^{(k)}
\right)^\top
\left(
e_\tau^{(k)}
+
r_\tau^{(k)}
\right),
\qquad
k=0,1,\ldots,K.
\]
Under the conditions of Lemma~\ref{lem:uniform-empirical-gram},
\(\mathbb P(\mathcal E_{\Sigma,\tau})\longrightarrow1\). On this event,
\begin{equation}
E_{\tau,s}
\lesssim
\sum_{k=1}^K
w_k
\left\|
g_\tau^{(k)}
\right\|_2
+
\frac{1}{\mu}
\left(
\left\|
g_\tau^{(0)}
\right\|_2
+
E_{\tau,0}
\right).
\label{eq:app-h9}
\end{equation}

Moreover, uniformly over \(\tau=1,\ldots,\Upsilon\),
\begin{equation}
\left\|
g_\tau^{(k)}
\right\|_2
=
O_p
\left(
\sqrt{\frac{q}{n_k}}
+
p^{-\kappa/d}
\right),
\qquad
k=0,1,\ldots,K.
\label{eq:app-h10}
\end{equation}
Consequently, if
\[
E_{\tau,0}
=
O_p
\left(
r_{\tau,0}
\right)
\]
for some deterministic rate \(r_{\tau,0}\), then
\begin{equation}
E_{\tau,s}
=
O_p
\left[
\sqrt{\frac{Kq}{n_s}}
+
p^{-\kappa/d}
+
\frac{1}{\mu}
\left\{
\sqrt{\frac{q}{n_0}}
+
p^{-\kappa/d}
+
r_{\tau,0}
\right\}
\right].
\label{eq:app-h11}
\end{equation}

In particular, if \(K=O(1)\), \(\mu=n_s/n_0\), and \(n_s\gtrsim n_0\), then
\begin{equation}
E_{\tau,s}
=
O_p
\left[
\sqrt{\frac{q}{n_s}}
+
p^{-\kappa/d}
+
\frac{n_0}{n_s}
r_{\tau,0}
\right].
\label{eq:app-h12}
\end{equation}
\end{lemma}

Lemma~\ref{lem:weighted-source-reconstruction} shows that the weighted source
estimation error consists primarily of the source regression rate
\(\sqrt{q/n_s}\), while the effect of the target estimation error is attenuated
by the factor \(1/\mu=n_0/n_s\). We first prove the two auxiliary lemmas and
then combine their conclusions directly to close the finite-iteration
recursion for \(d_\tau\).

\begin{proof}[Proof of Lemma~\ref{lem:weighted-bellman-discrepancy}]
Fix \(k\in[K]\). By the definition of the task-specific Bellman coefficients
in \eqref{eq:app-h2},
\[
\begin{aligned}
\beta_\tau^{(k)}
-
\beta_\tau^{(0)}
={}&
\beta_r^{(k)}
-
\beta_r^{(0)}
\\
&+
\gamma
\int_{\mathcal X}
V_{\tau-1}^{(k)}(x')
\left[
\beta_\rho^{(k)}(x')
-
\beta_\rho^{(0)}(x')
\right]
\,dx'
\\
&+
\gamma
\int_{\mathcal X}
\left[
V_{\tau-1}^{(k)}(x')
-
V_{\tau-1}^{(0)}(x')
\right]
\beta_\rho^{(0)}(x')
\,dx'.
\end{aligned}
\]

The first term is bounded directly by the reward component of \(h_k\). For
the second term, the uniform boundedness of the fitted value functions gives
\[
\begin{aligned}
&
\left\|
\int_{\mathcal X}
V_{\tau-1}^{(k)}(x')
\left[
\beta_\rho^{(k)}(x')
-
\beta_\rho^{(0)}(x')
\right]
\,dx'
\right\|_1
\\
&\qquad\le
B
\int_{\mathcal X}
\left\|
\beta_\rho^{(k)}(x')
-
\beta_\rho^{(0)}(x')
\right\|_1
\,dx'.
\end{aligned}
\]
Consequently, the first two terms are bounded by a constant multiple of
\(h_k\).

For the third term, define
\[
G_{\tau-1}^{(k)}(x')
=
V_{\tau-1}^{(k)}(x')
-
V_{\tau-1}^{(0)}(x').
\]
The transition-coefficient stability condition implies
\[
\left\|
\int_{\mathcal X}
G_{\tau-1}^{(k)}(x')
\beta_\rho^{(0)}(x')
\,dx'
\right\|_1
\le
C_\rho
\int_{\mathcal X}
\left|
G_{\tau-1}^{(k)}(x')
\right|
\,dx'.
\]

Using
\[
\left|
\max_{a'}u(a')
-
\max_{a'}v(a')
\right|
\le
\max_{a'}
|u(a')-v(a')|,
\]
we obtain
\[
\begin{aligned}
\left|
G_{\tau-1}^{(k)}(x')
\right|
&\le
\max_{a'\in\mathcal A}
\left|
\widehat Q_{\tau-1}^{(k)}(x',a')
-
\widehat Q_{\tau-1}^{(0)}(x',a')
\right|
\\
&=
\max_{a'\in\mathcal A}
\left|
\xi^\top(x',a')
\left(
\widehat\beta_{\tau-1}^{(k)}
-
\widehat\beta_{\tau-1}^{(0)}
\right)
\right|.
\end{aligned}
\]

By the second-moment condition on the sieve basis,
\[
\begin{aligned}
\int_{\mathcal X}
\left|
G_{\tau-1}^{(k)}(x')
\right|
\,dx'
\lesssim
\left\|
\widehat\beta_{\tau-1}^{(k)}
-
\widehat\beta_{\tau-1}^{(0)}
\right\|_2.
\end{aligned}
\]
Therefore,
\[
\left\|
\beta_\tau^{(k)}
-
\beta_\tau^{(0)}
\right\|_1
\lesssim
h_k
+
\gamma
\left\|
\widehat\beta_{\tau-1}^{(k)}
-
\widehat\beta_{\tau-1}^{(0)}
\right\|_2.
\]

Finally, since
\[
\delta_{\tau,\Delta}^{(0)}
=
\beta_\tau^{(0)}
-
\sum_{k=1}^K
w_k\beta_\tau^{(k)}
=
\sum_{k=1}^K
w_k
\left(
\beta_\tau^{(0)}
-
\beta_\tau^{(k)}
\right),
\]
the triangle inequality yields
\[
\begin{aligned}
\left\|
\delta_{\tau,\Delta}^{(0)}
\right\|_1
&\le
\sum_{k=1}^K
w_k
\left\|
\beta_\tau^{(0)}
-
\beta_\tau^{(k)}
\right\|_1
\\
&\lesssim
\sum_{k=1}^K
w_k h_k
+
\gamma
\sum_{k=1}^K
w_k
\left\|
\widehat\beta_{\tau-1}^{(k)}
-
\widehat\beta_{\tau-1}^{(0)}
\right\|_2
\\
&=
h_{\mathrm w}
+
\gamma d_{\tau-1}.
\end{aligned}
\]
This proves Lemma~\ref{lem:weighted-bellman-discrepancy}.
\end{proof}

\begin{proof}[Proof of Lemma~\ref{lem:weighted-source-reconstruction}]
For each task \(k=0,1,\ldots,K\), define
\[
u_\tau^{(k)}
=
\widehat\beta_\tau^{(k)}
-
\beta_\tau^{(k)}.
\]
Recall that
\[
g_\tau^{(k)}
=
\frac{1}{n_k}
\left(
Z_\tau^{(k)}
\right)^\top
\left(
e_\tau^{(k)}
+
r_\tau^{(k)}
\right).
\]

Let
\[
\widehat\delta_\tau
=
\widehat\beta_\tau^{(0)}
-
\sum_{k=1}^K
w_k\widehat\beta_\tau^{(k)}.
\]
By the first-order optimality conditions for the convex joint objective, there
exists
\[
\widehat z_\tau
\in
\partial
\left\|
\widehat\delta_\tau
\right\|_1
\]
such that the target and source KKT equations hold.

For the target coefficient, the KKT equation is
\[
-
\frac{1}{n_0}
\left(
Z_\tau^{(0)}
\right)^\top
\left(
y_\tau^{(0)}
-
Z_\tau^{(0)}
\widehat\beta_\tau^{(0)}
\right)
+
\lambda_\Delta
\widehat z_\tau
=
0.
\]
Using the regression representation
\[
y_\tau^{(0)}
=
Z_\tau^{(0)}
\beta_\tau^{(0)}
+
e_\tau^{(0)}
+
r_\tau^{(0)},
\]
we obtain
\[
\widehat\Sigma_\tau^{(0)}
u_\tau^{(0)}
-
g_\tau^{(0)}
+
\lambda_\Delta
\widehat z_\tau
=
0.
\]
Equivalently,
\begin{equation}
\lambda_\Delta
\widehat z_\tau
=
g_\tau^{(0)}
-
\widehat\Sigma_\tau^{(0)}
u_\tau^{(0)}.
\label{eq:app-h13}
\end{equation}

For the coefficient of source task \(k\), the derivative of the fusion
penalty contributes
\(-\lambda_\Delta w_k\widehat z_\tau\). Hence the corresponding KKT equation
is
\[
-
\mu w_k
\frac{1}{n_k}
\left(
Z_\tau^{(k)}
\right)^\top
\left(
y_\tau^{(k)}
-
Z_\tau^{(k)}
\widehat\beta_\tau^{(k)}
\right)
-
\lambda_\Delta
w_k\widehat z_\tau
=
0.
\]
Substituting
\[
y_\tau^{(k)}
=
Z_\tau^{(k)}
\beta_\tau^{(k)}
+
e_\tau^{(k)}
+
r_\tau^{(k)}
\]
gives
\[
\mu w_k
\left[
\widehat\Sigma_\tau^{(k)}
u_\tau^{(k)}
-
g_\tau^{(k)}
\right]
-
\lambda_\Delta
w_k\widehat z_\tau
=
0.
\]
Since \(w_k>0\),
\begin{equation}
\widehat\Sigma_\tau^{(k)}
u_\tau^{(k)}
=
g_\tau^{(k)}
+
\frac{\lambda_\Delta}{\mu}
\widehat z_\tau.
\label{eq:app-h14}
\end{equation}

Directly bounding \(\widehat z_\tau\) in \(\ell_2\) would introduce an
unnecessary factor \(\sqrt q\). Instead, substituting the target KKT identity
\eqref{eq:app-h13} into \eqref{eq:app-h14} yields
\[
\widehat\Sigma_\tau^{(k)}
u_\tau^{(k)}
=
g_\tau^{(k)}
+
\frac{1}{\mu}
\left[
g_\tau^{(0)}
-
\widehat\Sigma_\tau^{(0)}
u_\tau^{(0)}
\right].
\]
Therefore,
\begin{equation}
u_\tau^{(k)}
=
\left(
\widehat\Sigma_\tau^{(k)}
\right)^{-1}
\left\{
g_\tau^{(k)}
+
\frac{1}{\mu}
\left[
g_\tau^{(0)}
-
\widehat\Sigma_\tau^{(0)}
u_\tau^{(0)}
\right]
\right\}.
\label{eq:app-h15}
\end{equation}

On \(\mathcal E_{\Sigma,\tau}\),
\[
\left\|
\left(
\widehat\Sigma_\tau^{(k)}
\right)^{-1}
\right\|_{\mathrm{op}}
\le
c_{\mathrm G}^{-1},
\qquad
\left\|
\widehat\Sigma_\tau^{(0)}
\right\|_{\mathrm{op}}
\le
C_{\mathrm G}.
\]
It follows from \eqref{eq:app-h15} that
\[
\left\|
u_\tau^{(k)}
\right\|_2
\lesssim
\left\|
g_\tau^{(k)}
\right\|_2
+
\frac{1}{\mu}
\left(
\left\|
g_\tau^{(0)}
\right\|_2
+
\left\|
u_\tau^{(0)}
\right\|_2
\right).
\]
Multiplying by \(w_k\), summing over \(k\in[K]\), and using
\(\sum_{k=1}^K w_k=1\), we obtain
\[
E_{\tau,s}
\lesssim
\sum_{k=1}^K
w_k
\left\|
g_\tau^{(k)}
\right\|_2
+
\frac{1}{\mu}
\left(
\left\|
g_\tau^{(0)}
\right\|_2
+
E_{\tau,0}
\right).
\]
This proves \eqref{eq:app-h9}.

It remains to establish the score bound \eqref{eq:app-h10}. Decompose
\[
g_\tau^{(k)}
=
g_{\tau,e}^{(k)}
+
g_{\tau,r}^{(k)},
\]
where
\[
g_{\tau,e}^{(k)}
=
\frac{1}{n_k}
\left(
Z_\tau^{(k)}
\right)^\top
e_\tau^{(k)}
\]
and
\[
g_{\tau,r}^{(k)}
=
\frac{1}{n_k}
\left(
Z_\tau^{(k)}
\right)^\top
r_\tau^{(k)}.
\]

We first control the stochastic component. Conditional on
\(\mathcal F_{\tau-1}\), the Bellman regression errors are mean-zero and have
uniformly bounded conditional second moments. Moreover, the cross terms
vanish by the martingale-difference property. Therefore,
\[
\begin{aligned}
\mathbb E
\left[
\left\|
g_{\tau,e}^{(k)}
\right\|_2^2
\mid
\mathcal F_{\tau-1}
\right]
&=
\frac{1}{n_k^2}
\sum_{(i,t)\in S_\tau^{(k)}}
\mathbb E
\left[
\left(e_{i,t}^{(k),\tau}\right)^2
\left\|
\xi
\left(
X_{i,t}^{(k)},A_{i,t}^{(k)}
\right)
\right\|_2^2
\mid
\mathcal F_{\tau-1}
\right]
\\
&\lesssim
\frac{q}{n_k}.
\end{aligned}
\]
Hence, by Markov's inequality,
\begin{equation}
\left\|
g_{\tau,e}^{(k)}
\right\|_2
=
O_p
\left(
\sqrt{\frac{q}{n_k}}
\right).
\label{eq:app-h16}
\end{equation}

For the sieve-approximation component, using
\(\|r_\tau^{(k)}\|_\infty\lesssim p^{-\kappa/d}\), we have
\begin{equation}
\begin{aligned}
\left\|
g_{\tau,r}^{(k)}
\right\|_2
&=
\left\|
\frac{1}{n_k}
\left(
Z_\tau^{(k)}
\right)^\top
r_\tau^{(k)}
\right\|_2
\\
&\le
\frac{
\left\|
Z_\tau^{(k)}
\right\|_{\mathrm{op}}
}{
n_k
}
\left\|
r_\tau^{(k)}
\right\|_2
\\
&=
\sqrt{
\lambda_{\max}
\left(
\widehat\Sigma_\tau^{(k)}
\right)
}
\frac{
\left\|
r_\tau^{(k)}
\right\|_2
}{
\sqrt{n_k}
}
\\
&\lesssim
p^{-\kappa/d}.
\end{aligned}
\label{eq:app-h17}
\end{equation}
Combining \eqref{eq:app-h16} and \eqref{eq:app-h17} proves
\[
\left\|
g_\tau^{(k)}
\right\|_2
=
O_p
\left(
\sqrt{\frac{q}{n_k}}
+
p^{-\kappa/d}
\right),
\]
which is \eqref{eq:app-h10}.

Next,
\[
\begin{aligned}
\sum_{k=1}^K
w_k
\sqrt{\frac{q}{n_k}}
&=
\frac{\sqrt q}{n_s}
\sum_{k=1}^K
\sqrt{n_k}
\\
&\le
\frac{\sqrt q}{n_s}
\sqrt{
K
\sum_{k=1}^K n_k
}
\\
&=
\sqrt{\frac{Kq}{n_s}},
\end{aligned}
\]
where the inequality follows from Cauchy--Schwarz. Thus, whenever
\(E_{\tau,0}=O_p(r_{\tau,0})\), \eqref{eq:app-h9} and \eqref{eq:app-h10} imply
\[
E_{\tau,s}
=
O_p
\left[
\sqrt{\frac{Kq}{n_s}}
+
p^{-\kappa/d}
+
\frac{1}{\mu}
\left\{
\sqrt{\frac{q}{n_0}}
+
p^{-\kappa/d}
+
r_{\tau,0}
\right\}
\right],
\]
which proves \eqref{eq:app-h11}.

Finally, suppose \(K=O(1)\), \(\mu=n_s/n_0\), and \(n_s\gtrsim n_0\). Then
\[
\sqrt{\frac{Kq}{n_s}}
\asymp
\sqrt{\frac{q}{n_s}},
\]
and
\[
\frac{1}{\mu}
\sqrt{\frac{q}{n_0}}
=
\frac{n_0}{n_s}
\sqrt{\frac{q}{n_0}}
=
\sqrt{\frac{q}{n_s}}
\sqrt{\frac{n_0}{n_s}}
\lesssim
\sqrt{\frac{q}{n_s}}.
\]
Moreover,
\[
\frac{1}{\mu}
p^{-\kappa/d}
\lesssim
p^{-\kappa/d}.
\]
It follows that
\[
E_{\tau,s}
=
O_p
\left[
\sqrt{\frac{q}{n_s}}
+
p^{-\kappa/d}
+
\frac{n_0}{n_s}
r_{\tau,0}
\right].
\]
This proves \eqref{eq:app-h12} and completes the proof.
\end{proof}

Define
\[
H_\tau
=
\left\|
\delta_{\tau,\Delta}^{(0)}
\right\|_1
\]
and, for \(x\ge0\), define
\[
\Delta_s(x)
=
\sqrt{s}\,\overline\lambda_\Delta
\wedge
\sqrt{x\overline\lambda_\Delta}
\wedge
x.
\]
Also let
\[
r_n
=
\sqrt{\frac{q}{n_{\mathrm{all}}}}
+
\sqrt{\frac{q}{n_s}}
+
p^{-\kappa/d}.
\]

The proof of Lemma~\ref{lem:one-step-target-coefficient} is conditional on
the preceding fitted iterates and uses \(h_\Delta\) only through the
iteration-specific bound
\[
\left\|
\delta_{\tau,\Delta}^{(0)}
\right\|_1
\le
h_\Delta.
\]
Therefore, the same argument applied with \(H_\tau\) in place of
\(h_\Delta\) gives
\begin{equation}
E_{\tau,0}
=
O_p
\left[
\sqrt{\frac{q}{n_{\mathrm{all}}}}
+
p^{-\kappa/d}
+
\Delta_s(H_\tau)
\right].
\label{eq:app-h18}
\end{equation}

We now close the recursion by induction. Suppose that all tasks use a common
initialization,
\[
\widehat\beta_0^{(k)}
=
\widehat\beta_0^{(0)},
\qquad
k\in[K],
\]
so that
\[
d_0=0.
\]

Suppose, for some \(\tau\ge1\), that
\begin{equation}
d_{\tau-1}
=
O_p
\left(
h_{\mathrm w}+r_n
\right).
\end{equation}
Lemma~\ref{lem:weighted-bellman-discrepancy} then gives
\begin{equation}
H_\tau
\lesssim
h_{\mathrm w}
+
\gamma d_{\tau-1}
=
O_p
\left(
h_{\mathrm w}+r_n
\right).
\label{eq:app-h20}
\end{equation}

The function \(\Delta_s(\cdot)\) is nondecreasing and satisfies, up to
absolute constants,
\begin{equation}
\Delta_s(x+y)
\lesssim
\Delta_s(x)+y,
\qquad
x,y\ge0.
\end{equation}
Indeed, \(\Delta_s(x)\le x\), and its three-branch form is nondecreasing and
subadditive. Consequently, \eqref{eq:app-h20} implies
\begin{equation}
\Delta_s(H_\tau)
=
O_p
\left[
\Delta_s(h_{\mathrm w})
+
r_n
\right].
\label{eq:app-h22}
\end{equation}

Combining \eqref{eq:app-h18} and \eqref{eq:app-h22}, we obtain
\begin{equation}
E_{\tau,0}
=
O_p
\left[
r_n
+
\Delta_s(h_{\mathrm w})
\right].
\label{eq:app-h23}
\end{equation}
Since
\[
\Delta_s(h_{\mathrm w})
\le
h_{\mathrm w},
\]
it also follows that
\begin{equation}
E_{\tau,0}
=
O_p
\left(
r_n+h_{\mathrm w}
\right).
\label{eq:app-h24}
\end{equation}

Under \(K=O(1)\), \(\mu=n_s/n_0\), and \(n_s\gtrsim n_0\),
Lemma~\ref{lem:weighted-source-reconstruction} and \eqref{eq:app-h24} yield
\begin{equation}
\begin{aligned}
E_{\tau,s}
&=
O_p
\left[
\sqrt{\frac{q}{n_s}}
+
p^{-\kappa/d}
+
\frac{n_0}{n_s}
\left(
r_n+h_{\mathrm w}
\right)
\right]
\\
&=
O_p
\left(
r_n+h_{\mathrm w}
\right).
\end{aligned}
\label{eq:app-h25}
\end{equation}

Substituting \eqref{eq:app-h20}, \eqref{eq:app-h24}, and \eqref{eq:app-h25} into \eqref{eq:app-h8} gives
\[
d_\tau
=
O_p
\left(
h_{\mathrm w}+r_n
\right).
\]
This verifies the induction step. Since \(d_0=0\) and the number of FQI
iterations \(\Upsilon\) is fixed,
\begin{equation}
\max_{0\le\tau\le\Upsilon}
d_\tau
=
O_p
\left(
h_{\mathrm w}+r_n
\right).
\end{equation}

Applying \eqref{eq:app-h20}--\eqref{eq:app-h23} uniformly over the finitely many iterations gives
\begin{equation}
\max_{1\le\tau\le\Upsilon}
E_{\tau,0}
=
O_p
\left[
r_n
+
\Delta_s(h_{\mathrm w})
\right].
\label{eq:app-h27}
\end{equation}

In the source-rich regime \(n_s\gtrsim n_0\),
\[
n_{\mathrm{all}}
=
n_0+n_s
\asymp
n_s,
\]
and hence
\[
r_n
\asymp
\sqrt{\frac{q}{n_{\mathrm{all}}}}
+
p^{-\kappa/d}.
\]
Therefore,
\begin{equation}
\boxed{
\max_{1\le\tau\le\Upsilon}
\left\|
\widehat\beta_\tau^{(0)}
-
\beta_\tau^{(0)}
\right\|_2
=
O_p
\left[
\sqrt{\frac{q}{n_{\mathrm{all}}}}
+
p^{-\kappa/d}
+
\Delta_s(h_{\mathrm w})
\right].
}
\end{equation}

Finally, the same second-moment argument used in the homogeneous-transition
analysis gives, for the target-reachable class \(\mathfrak D_0\),
\[
\sup_{\nu\in\mathfrak D_0}\mathbb E_\nu
\left|
\mathcal T^{(0)}
\widehat Q_{\tau-1}^{(0)}
-
\widehat Q_\tau^{(0)}
\right|
\lesssim
p^{-\kappa/d}
+
E_{\tau,0}.
\]
Combining this inequality with \eqref{eq:app-h1}, \eqref{eq:app-h27}, and the geometric-series bound
gives
\begin{equation}
\sup_{\nu\in\mathfrak D_0}\mathbb E_\nu
\left|
Q^{*,(0)}
-
\widehat Q_\Upsilon^{(0)}
\right|
=
O_p
\left\{
\frac{1}{1-\gamma}
\left[
\sqrt{\frac{q}{n_{\mathrm{all}}}}
+
p^{-\kappa/d}
+
\Delta_s(h_{\mathrm w})
\right]
+
\frac{
\gamma^\Upsilon R_{\max}
}{
1-\gamma
}
\right\}.
\end{equation}

Applying the standard greedy-policy performance inequality yields
\begin{equation}
\begin{aligned}
v^{\pi^{*,(0)},(0)}
-
v^{\widehat\pi_\Upsilon^{(0)},(0)}
=
O_p\Bigg(
&
\frac{1}{(1-\gamma)^2}
\left[
\sqrt{\frac{q}{n_{\mathrm{all}}}}
+
p^{-\kappa/d}
+
\Delta_s(h_{\mathrm w})
\right]
\\
&+
\frac{
\gamma^\Upsilon R_{\max}
}{
(1-\gamma)^2
}
\Bigg).
\end{aligned}
\end{equation}
This completes the heterogeneous-transition analysis.

\end{document}